\documentclass[letterpaper]{article} % DO NOT CHANGE THIS
\usepackage{aaai25}  % DO NOT CHANGE THIS
\usepackage{times}  % DO NOT CHANGE THIS
\usepackage{helvet}  % DO NOT CHANGE THIS
\usepackage{courier}  % DO NOT CHANGE THIS
\usepackage[hyphens]{url}  % DO NOT CHANGE THIS
\usepackage{graphicx} % DO NOT CHANGE THIS
\usepackage{natbib}  % DO NOT CHANGE THIS AND DO NOT ADD ANY OPTIONS TO IT
\usepackage{caption} % DO NOT CHANGE THIS AND DO NOT ADD ANY OPTIONS TO IT
\usepackage{algorithm}
\usepackage{algorithmic}
\usepackage{multirow}
\usepackage{diagbox}
\usepackage{amsmath} 
\usepackage{amsthm}
\usepackage{amsfonts}       % blackboard math symbols
\usepackage{placeins}

\newtheorem{proposition}{Proposition}
\newtheorem{technique}{Technique}

\usepackage{newfloat}
\usepackage{listings}
\DeclareCaptionStyle{ruled}{labelfont=normalfont,labelsep=colon,strut=off} % DO NOT CHANGE THIS
\floatstyle{ruled}
\newfloat{listing}{tb}{lst}{}
\floatname{listing}{Listing}
\title{Supervised Score-Based Modeling by Gradient Boosting}
\author {
    Changyuan Zhao\textsuperscript{\rm 1,\rm 2},
    Hongyang Du\textsuperscript{\rm 3}\thanks{Corresponding Authors},
    Guangyuan Liu\textsuperscript{\rm 1, \rm 4},
    Dusit Niyato\textsuperscript{\rm 1}
}
\affiliations {
    \textsuperscript{\rm 1}College of Computing
and Data Science, Nanyang Technological University\\
\textsuperscript{\rm 2}CNRS@CREATE, 1 Create Way, \# 08-01 Create Tower, Singapore 138602\\
    \textsuperscript{\rm 3}Department of Electrical and Electronic Engineering, University of Hong Kong\\
    \textsuperscript{\rm 4}The Energy Research Institute @ NTU, Interdisciplinary
Graduate Program\\
    zhao0441@e.ntu.edu.sg, duhy@eee.hku.hk, liug0022@e.ntu.edu.sg, dniyato@ntu.edu.sg
}

\usepackage{bibentry}
\begin{document}

\maketitle

\begin{abstract}
Score-based generative models can effectively learn the distribution of data by estimating the gradient of the distribution. Due to the multi-step denoising characteristic, researchers have recently considered combining score-based generative models with the gradient boosting algorithm, a multi-step supervised learning algorithm, to solve supervised learning tasks. However, existing generative model algorithms are often limited by the stochastic nature of the models and the long inference time, impacting prediction performances. Therefore, we propose a Supervised Score-based Model (SSM), which can be viewed as a gradient boosting algorithm combining score matching. We provide a theoretical analysis of learning and sampling for SSM to balance inference time and prediction accuracy. Via the ablation experiment in selected examples, we demonstrate the outstanding performances of the proposed techniques. Additionally, we compare our model with other probabilistic models, including Natural Gradient Boosting (NGboost), Classification and Regression Diffusion Models (CARD), Diffusion Boosted Trees (DBT), and non-probabilistic GBM models. The experimental results show that our model outperforms existing models in both accuracy and inference time.
\end{abstract}


\section{Introduction}

Generative artificial intelligence (GAI) models aim to create highly realistic simulations or reproductions of data by learning from the characteristics of the original data. 
Recently, score-based models (diffusion models) have become the most powerful families of GAI due to their excellent generation performance \cite{yang2023diffusion, dhariwal2021diffusion}. 
In these models, a series of noises is first added to the original data through a forward diffusion process. The model then learns the added noise and continuously denoises it to generate new data that follows the distribution of the original dataset. 
At present, many researchers have applied score-based models to generation tasks in many fields, including image generation, audio generation, video generation, etc., all with excellent performance \cite{rombach2021highresolution, kong2020diffwave, ho2022video}. 
% These works demonstrate the powerful learning ability of score-based models for complex distributions in many fields.

Supervised learning is another important research area in machine learning different from generative learning.
In supervised learning, a model is trained on labeled data with the goal of learning a mapping from inputs to outputs, allowing it to predict the output for unseen data accurately \cite{cunningham2008supervised}. 
Regression and classification problems are two basic types of supervised learning tasks, which aim to train a model for prediction. For regression problems, this prediction result is the target output corresponding to the input data, while for classification problems, it is the accurate label. The learning performance of this data relationship is closely related to the learning ability of the sophisticated structure of neural network models \cite{he2016deep}.
Particularly, 
% the Gradient Boosting Machine (GBM) is a supervised learning model that continuously corrects errors through iteration. 
% GBM model combines several weak learners into a strong learner
%  The sequential weak learners reduce errors from previous iterations to obtain accurate predictions, especially with large and complex datasets 
the Gradient Boosting Machine (GBM) is a supervised learning model that combines several weak learners into a strong learner \cite{bentejac2021comparative}.
The sequential weak learners reduce errors from previous iterations to obtain accurate predictions, especially with large and complex datasets. 
Recently, GBM algorithms such as XGBoost \cite{chen2016xgboost} and LightGBM \cite{ke2017lightgbm} have demonstrated state-of-the-art performance in tabular competitions on platforms, including Kaggle.

% iteration. It combines several weak learners into a strong learner. The sequential weak learners reduce errors from previous iterations to obtain accurate predictions, especially with large and complex datasets.

Due to the multi-step denoising characteristic,
researchers have recently considered combining score-based generative models with the gradient boosting algorithm to solve supervised learning tasks \cite{han2024diffusion, beltran2024treeffuser}.
% Due to the ability of generative models to capture data features, many researchers have studied the applications in supervised learning, including image classification and regression \cite{wu2024medsegdiff, han2022card}. 
% Additionally, 
% score-based models' continuous estimation of noise contributes significantly to addressing uncertainty estimation, a critical task in supervised learning \cite{zhou2022survey}. 
However, these previous methods are mainly based on diffusion models, whose random denoising process may lead to unstable prediction results and reduce the robustness. Moreover, diffusion models often require longer denoising steps, increasing model inference time. If accelerated algorithms such as DDIM \cite{song2020denoising} are used, the quality of the generated results will usually decrease, resulting in lower prediction accuracy.

Therefore, in this paper, our goal is to propose a score-based gradient boosting model for supervised learning to enable quick and accurate inference. To achieve this goal, we consider the score-based generative models via denoising score matching \cite{song2019generative} since its inference process does not have a specific number of denoising steps. Based on this, we propose \textbf{Supervised Score-based Model (SSM)}, a score-based model by gradient boosting.

We summarize our main contributions as follows: 
1) We utilize noise-free Langevin Dynamics to establish a connection between score-based generative models and GBM algorithms.
2) we present SSM to predict the maximum log-likelihood estimation point, which estimates errors between the target and input data through a score network. 3) For the training process and inference process, we provide a theoretical analysis to balance inference time and prediction accuracy. 4) Experiments on regression and classification tasks show that SSM can achieve better performance than existing methods and significantly shorten the inference time.

\section{Related Work}
Score-based generative models have demonstrated powerful learning capabilities for complex data distribution in many fields \cite{croitoru2023diffusion, du2023beyond, liu2024generative}. There are generally two categories of such models: diffusion-based and score matching-based models \cite{yang2023diffusion}. Diffusion-based models consist of a forward diffusion process, which adds Gaussian noises into the original data, making it converge to the standard normal distribution, and a denoising process, which removes the added noises from a random variable following normal distribution \cite{ho2020denoising, nichol2021improved}. The score matching-based model can be regarded as an energy-based model \cite{song2021train}, which aims to learn explicit probability distributions of data by a score function. Denoising score matching is a state-of-the-art approach for score estimation, which also estimates the score function from the disturbed dataset and then denoises through Langevin dynamics \cite{song2019generative}. From a general perspective, these two types of models can also be viewed as the discretization of stochastic differential equations \cite{song2020score}.

The GBM is a powerful ensemble learning technique that builds models sequentially, with each new model aiming to correct the errors of its predecessors \cite{friedman2001greedy}. By combining the predictions of multiple weak learners, GBM creates a strong predictive model that can handle complex data patterns and interactions \cite{natekin2013gradient}. 
XGBoost, LightGBM, and NGBoost are typical decision trees-based gradient boosting techniques that extend the foundational principles of GBM to enhance efficiency and performance. XGBoost and LightGBM focus on computational speed and scalability \cite{chen2016xgboost, ke2017lightgbm}, while NGBoost introduces probabilistic predictions, offering a novel approach to uncertainty estimation in gradient boosting models \cite{duan2020ngboost}. 
% These advancements make GBM well-suited for complex supervised learning tasks, offering improved accuracy, speed, and interpretability in predictive modeling.

% NGBoost \cite{duan2020ngboost},

%  XGBoost \cite{chen2016xgboost} and LightGBM \cite{ke2017lightgbm} 

% The method is widely used in various machine learning tasks due to its ability to optimize predictive accuracy and robustness, making it a popular choice for both classification and regression problems.

Recently, 
via setting input data in supervised learning as conditions, conditional score-based models have been utilized to 
solve supervised learning problems \cite{amit2021segdiff, rahman2023ambiguous, beltran2024treeffuser}.
\cite{zimmermann2021score} proposed a score-based generative classifier for classification tasks. This model predicts classification results through maximum likelihood estimation of different label generation distributions and target values instead of direct prediction.
In \cite{han2022card}, the authors proposed CARD, a score-based model for classification and regression. They emphasized its ability to solve uncertainty in supervised learning and compared its performance with existing Bayesian neural networks \cite{blundell2015weight, tomczak2021collapsed}.
The DBT model, a diffusion boosting paradigm, connects the denoising diffusion generative model and GBM via decision trees. Through experiments on real-world regression tasks, this approach demonstrates the potential of integrating GBM with score-based generative models \cite{han2024diffusion}.

% Conditional score-based models are generative models that can utilize extra input as the condition to guide the denoising process for generating results that satisfy certain constraints \cite{zhang2023adding}. Leveraging the input condition, conditional score-based models are capable of many multi-modal tasks, including text-to-image, text-to-video
% super-resolution, etc, \cite{zhang2023text, kawar2022denoising, khachatryan2023text2video}. Additionally, when setting input data in supervised learning as conditions, conditional score-based models can naturally solve supervised learning problems \cite{amit2021segdiff, rahman2023ambiguous}.
% \cite{zimmermann2021score} proposed a score-based generative classifier for classification tasks. This model predicts classification results through maximum likelihood estimation of different label generation distributions and target values instead of direct prediction.
% In \cite{han2022card}, the authors proposed CARD, a score-based model for classification and regression. They emphasized its ability to solve uncertainty in supervised learning and compared its performance with existing Bayesian neural networks \cite{blundell2015weight, tomczak2021collapsed}.

% \subsection{Distribution Estimation for Regression Problem}

\section{Background}

\subsection{Gradient Boosting Machine}

The GBM is a supervised learning framework that combines several weak learners into strong learners in an iterative way \cite{friedman2001greedy}.
In GBM, a boosted model $F(x)$ is a weighted linear combination of $m$ weak learners, which can be formulated as
\begin{equation}
\label{eq:BGM}
    F(x) = F_m(x) = F_0(x) + \sum_{k=1}^{m} \alpha_k h_k(x),
\end{equation}
where $F_0(x)$ is the initial prediction, $\alpha_k$ is the weight coefficient, and $h_k(x)$ represents weak learner. 
Let $l(y, F(x))$ represent a measure of data error at the observation $(y, x)$ for the differentiable loss function $l$.
To train a GBM, the goal is to explore a function $F$ that minimizes the expected loss $L(F(x)) = E_{(x,y)}l(y,F(x))$ where the expectation is taken over the input dataset of $(y, x)$.
Since the loss function $l$ is differentiable, the optimal solution $F^*(x)$ can be represented as
\begin{equation}
\label{eq:optimal1}
    F^*(x) = F_0(x) + \sum_{k=1}^{m} \alpha_k \cdot(-g_k(x)),
\end{equation}
where $g_k(x) = \nabla_{F_{k-1}(x)} L(F_{k-1}(x))$ is the gradient at optimization step $k$. Therefore, weak learners are trained to estimate the negative gradient term to approximate the optimal solution.
Given a trained GBM and input, the predicted result can be obtained via the iterative equation Eq. \eqref{eq:BGM}.

%  A primary goal of machine learning is
% to obtain a function f that minimizes the expected loss EP (`(y, f(x))) where the expectation is
% taken over the unknown distribution of (y, x) (denoted by P)

% To train a GBM, the loss function is $l(F(x)) = E_{x,y} L(y, F(x))$ and $h_k(x) = -\nabla l(F_{k-1}(x))$ 
% is the gradient at optimization step $k$. 

% When we set the loss function $l(F(x))$ of the boosted model as the negative log-likelihood estimation, i.e., $-log p(F(x))$, 
% the boosted model can be formulated as 
% $$F_{k}(x) = F_{k-1}(x) + \beta_{k} \cdot \nabla log p(F_{k-1}(x))$$
% and the prediction of $y$ given any $x$ can be obtained by 
% $$\widehat{y} = F(x) = F_0(x) + \sum_{k=1}^{m} \beta_{k} \cdot \nabla log p(F_{k-1}(x)).$$
% The formulated boosted model has a similar structure to Langevin dynamics without noise terms.
% Since our motivation is to integrate the denoising process with the gradient boosting process, we employ noise-free Langevin dynamics in our Supervised Score-based Model (SSM).

\subsection{Langevin Dynamics}

Langevin dynamics is a mathematical model that describes the dynamics of molecular systems. 
For a continuously differentiable probability distribution $p(x)$ where $x\in\mathbb{R}^d$, Langevin dynamics uses the score function, i.e., log-likelihood estimate $\nabla_x \log p(x)$, iteratively to obtain samples that fit the distribution $p(x)$ \cite{welling2011bayesian}. Given a certain step size $\epsilon > 0$ and any prior distribution $\pi(x)$, the Langevin dynamics can be expressed as
\begin{equation}
\label{Langevin}
x_t = x_{t-1} + \frac{\epsilon}{2} \nabla_x \log p(x_{t-1}) + \sqrt{\epsilon} z_t,
\end{equation}
where $t\geq 0$, $x_0 \sim \pi(x)$, and $z_t \sim \mathcal{N}(0,1)$. As a Markov chain Monte Carlo (MCMC) technique \cite{robert1999monte}, the Langevin equation takes gradient steps based on the score function and also injects Gaussian noise to capture the distribution 
$p(x)$ instead of the maximum point of the log-likelihood estimate $\log p(x)$. When $\epsilon\rightarrow 0$ and $t\rightarrow \infty$, the distribution of $x_t$ will converge to the target distribution $p(x)$ under some regularity conditions \cite{roberts1996exponential}.

\subsection{Denoising Score Matching for Score Estimation}

Since the Langevin equation in Eq. \eqref{Langevin} only relies on the score function $\log p(x)$, estimating the score of the target distribution is a crucial step. Recent studies \cite{hyvarinen2005estimation, song2020sliced, vincent2011connection} have explored the use of the score network $s_\theta:\mathbb{R}^d\rightarrow\mathbb{R}^d$, a neural network parameterized by $\theta$, to estimate the score function.

% Following \cite{hyvarinen2005estimation}, we can train the score network by minimizing the loss defined as follows:
% \begin{equation}
% \label{SMloss}
% J(\theta) = \mathbb{E}_{p(x)} [\Tr(\nabla_x s_\theta (x)) + \frac{1}{2} \Vert s_\theta(x)\Vert_2^2],
% \end{equation}
% where $\nabla_x s_\theta (x)$ represents the Jacobian matrix of $s_\theta(x)$. However, calculating the Jacobian matrix of a neural network is complicated especially for large scale networks. To overcome it, sliced score matching \cite{song2020sliced}
% uses random projections to approximate $\Tr(\nabla_x s_\theta (x))$, computed by forward mode auto-differentiation, which still requires a relatively long computation time. 

\subsubsection{Denoising Score Matching}

% \textbf{Denoising Score Matching}~~~ 
Denoising score matching \cite{vincent2011connection} is a method 
% that completely avoids the calculation of $\Tr(\nabla_x s_\theta (x))$ 
to estimate the score function
by adding noise to the original data and then estimating the perturbed data distribution.
Specifically, perturbing the data with a specified noise can represent 
the perturbed distribution for point $x$ as $q(\widetilde{x}|x)$. Then, the entire perturbed distribution is $q(\widetilde{x}) = \int q(\widetilde{x}|x)p(x)dx$. To estimate the perturbed distribution, the loss function can be represented as 
% in Eq. \eqref{SMloss} can be converted into,
\begin{equation}
\label{DSMloss}
    J'(\theta) = \frac{1}{2} \mathbb{E}_{q(\widetilde{x}|x)p(x)}[\Vert s_{\theta} (\widetilde{x}) - \nabla_{\widetilde{x}}\log q(\widetilde{x}|x)\Vert_2^2].
\end{equation}
When the added noise is small enough, e.g., $q(x) \approx p(x)$, the optimal score network $s_{\theta^{*}}(x)$, where $\theta^{*}=\arg\min_\theta J'(\theta)$, can represent the score of original data distribution, i.e., $s_{\theta^{*}}(x) = \nabla_x \log q(x)\approx \nabla_x \log p(x)$ \cite{vincent2011connection}. 
However, denoising score matching may not accurately estimate the score in areas with low data density due to insufficient training data \cite{song2019generative}. This issue can be solved by adding larger noise, but it will not satisfy the condition that the added noise is sufficiently small, which can lead to another kind of error.

\subsubsection{Noise Conditional Score Network}

% \textbf{Noise Conditional Score Network}~~~
Noise Conditional Score Network (NCSN) is a well-designed score network that uses various levels of noise while simultaneously estimating scores using a single neural network \cite{song2019generative}. In NCSN, the noises are a series of normal distributions whose variances $\{\sigma_{i}\}_{i=1}^L$ are a positive geometric sequence, i.e., $\frac{\sigma_1}{\sigma_2} = \cdots = \frac{\sigma_{L-1}}{\sigma_L}>1$. Based on the noise levels $\{\sigma_{i}\}_{i=1}^L$, the perturbed data distribution is $q_{\sigma}(\widetilde{x}) = \int p(x)\mathcal{N}(\widetilde{x}|x, \sigma^2 I)dx$. NCSN aims to jointly estimate the scores of all perturbed data distribution, i.e., $\forall \sigma\in \{\sigma_{i}\}_{i=1}^L: s_\theta (x, \sigma)\approx \nabla_x \log q_\sigma (x)$, where $s_\theta (x,\sigma)\in\mathbb{R}^d$. For a specific noise level $\sigma$, the objective in Eq. \eqref{DSMloss} can be transferred into,
\begin{equation}
\label{NCSNloss1}
    l(\theta;\sigma) = \frac{1}{2}\mathbb{E}_{p(x)} \mathbb{E}_{\widetilde{x}\sim \mathcal{N}(x,\sigma^2 I)}[\Vert s_{\theta} (\widetilde{x}, \sigma) + \frac{\widetilde{x}-x}{\sigma^2}\Vert_2^2].
\end{equation}

Then for all noise levels $\{\sigma_{i}\}_{i=1}^L$, the loss is,
\begin{equation}
\label{NCSNloss2}
    \mathcal{L}(\theta;\{\sigma_{i}\}_{i=1}^L) = \frac{1}{L}\Sigma_{i=1}^L \lambda(\sigma_i) l(\theta; \sigma_i),
\end{equation}
where $\lambda(\sigma_i)$ is a coefficient depending on $\sigma_i$.
% , usually taken as $\sigma_i^2$.
After being trained, NCSN generates samples using annealed Langevin dynamics \cite{song2019generative} where the denoising process occurs level by level according to the noise level.
\section{Supervised Score-based
Model via Denoising Score Matching and Gradient Boosting}

% \subsection{Problem Statement and Model Definition}

In supervised learning, the training set $\mathcal{D}$ comprises multiple input-target pairs, i.e., $\mathcal{D} = \{(x_i,y_i)\}_{i=1}^N$, where $x_i \in \mathbb{R}^m$ and $y_i \in \mathbb{R}^d$. The model $f:\mathbb{R}^m\rightarrow\mathbb{R}^d$ trained on this training set aims to predict the response variable $y$ given an input variable $x$, i.e., $f(x) \approx y$.
For a regression problem, the response variable $y$ is a continuous variable, whereas it is a categorical variable for classification.

For supervised learning problems, given an input $x_i$, unlike estimating the distribution of output, we pay more attention to predicting the value of $y_i$, or a special single-point distribution, i.e. $p(y) = \mathbb{I}_{y_i}(y)$, where $\mathbb{I}(\cdot)$ is an indicator function. 
It is meaningless to estimate the score function of a single point distribution since its score function is only defined at value $y_i$. To address this issue, denoising score matching provides a feasible method via perturbing the single point distribution with noises. Specifically, when adding a Gaussian noise with mean 0 and variance $\sigma$, the perturbed distribution will become a normal distribution with mean $y_i$ and variance $\sigma$, i.e., 
$q(\widetilde{y}) = \int p(y) \mathcal{N}(\widetilde{y}|y, \sigma^2 I) dy = \mathcal{N}(y_i, \sigma^2)$, whose score function is well defined. 
 % When $\sigma$ is small enough, the perturbed distribution $q(\widetilde{y})$ 
Therefore, to estimate the value or the single point distribution of $y_i$, we aim to train an NCSN to estimate the score function of input-target pairs $(x_i,y_i)\in \mathcal{D}$ which uses an input variable $x_i$ as another 
condition, i.e., $s_\theta(y,\sigma_j,x_i)\in\mathbb{R}^d$, where $\sigma_j \in \{\sigma_{j}\}_{j=1}^L$. For a given $(x_i,y_i)\in \mathcal{D}$, $s_\theta(y,\sigma_j,x_i)$ denotes the score function of $y_i$, which is the score network of our framework SSM.

Accordingly, the objective in our framework is,
\begin{equation}
    l(\theta;\sigma, x,y) = \frac{1}{2}\mathbb{E}_{p(y)} \mathbb{E}_{\widetilde{y}\sim \mathcal{N}(y,\sigma^2 I)}[\Vert s_{\theta} (\widetilde{y}, \sigma, x) + \frac{\widetilde{y}-y}{\sigma^2}\Vert_2^2].
\end{equation}

Then for all noise levels $\{\sigma_{i}\}_{i=1}^L$ and dataset $\mathcal{D}$, the loss is,
\begin{equation}
\label{myloss}
    \mathcal{L}(\theta;\{\sigma_{i}\}_{i=1}^L) = \frac{1}{L}\Sigma_{i=1}^L \lambda(\sigma_i)\mathbb{E}_{(x,y)\in \mathcal{D}} l(\theta; \sigma_i, x,y).
\end{equation}

% To use denoising score matching in supervised learning, we aim to train an NCSN which uses an input variable $x$ as another condition, i.e., $s_\theta(y,\sigma_j,x_i)\in\mathbb{R}^d$, where $x_i \in $

% and produce samples following annealed Langevin dynamics. 

Due to the multi-step denoising characteristic of score-based generative models and GBM models, analyzing the relationship between the two is more helpful in deploying score-based generative models in supervised learning tasks.
Additionally, the choice of model parameters greatly affects the performance of the score-based generative model and the GBM model \cite{bentejac2021comparative}.
As discussed in \cite{song2020improved}, we need to design many parameters to ensure the effectiveness of training and inference, 
including (i) the choices of noise scales $\{\sigma_{i}\}_{i=1}^L$; (ii) the step size $\epsilon$ in Langevin dynamics; (iii) the inference steps $t$ in Langevin equation. 
% Additionally, 
% the strong generative diversity of the score-based generative model comes from the randomness of the denoising process \cite{yang2023diffusion}, i.e., the uncertain term $z_t$ in the Langevin equation, which may result in instability for supervised learning. 
% Additionally, the uncertain term $z_t$ in the Langevin equation (Eq. \eqref{Langevin}) will result in instability, which also needs to be considered.
% Specifical, we 
Therefore, in the following, we provide theoretical analysis to ensure the performance of the SSM on regression and classification problems.

\subsection{Gradient Boosting}

% Firstly, we consider adjust the Annealed Langevin dynamics (Alg. \ref{ald}) into a supervised version.

% It is well known that the strong generative diversity of the score-based generative model comes from the randomness of the denoising process \cite{yang2023diffusion}. 

Firstly, we consider the connection between SSM and GBM. For a given input, the denoising score match of SSM provides a perturbed
distribution in the solution space whose maximum log-likelihood estimation point is the target value.
% Thus, we can predict the output via the maximum log-likelihood estimation point
% Via the maximum log-likelihood estimation point we 
To predict the maximum log-likelihood estimation point, we can set the loss function $l$ in GBM as the negative log-likelihood estimation, i.e., $-log~q(F(x))$, where $q(\widetilde{x}|x)$ is the perturbed distribution. Thus, the optimal solution in Eq. \eqref{eq:optimal1} will be the maximum log-likelihood estimation point of the perturbed distribution.
Particularly, the iterative equation of GBM in Eq. \eqref{eq:BGM} can be converted to a noise-free Langevin equation with the above loss function, i.e.,
\begin{equation}
    F_{k}(x) = F_{k-1}(x) + \beta_{k} \cdot \nabla log~q(F_{k-1}(x)).
\end{equation}
Moreover, we plan to train one conditional score network with different noise levels as the conditions,
similar to the uniform score network in the score-based generative model.
The different noise level conditions represent various weak learners in GBM.
% Firstly, we consider the uncertain term $z_t$ in Langevin dynamics. 
% According to \cite{welling2011bayesian}, injecting Gaussian noise aims to simulate the whole distribution instead of the maximum point of the log-likelihood estimation. 
% However, for the single point distribution or the perturbed distribution in our supervised learning task, the maximum point is the target value. Therefore, to obtain estimates of points rather than distributions, we modify the Langevin equation by removing the uncertain term.
Specifically, given an input $x_I$ and a noise level $\sigma_j$, and a trained score network $s_\theta(y,\sigma_j,x_I)$, the converted noise-free Langevin equation is
\begin{equation}
\label{Langevin1}
y_t = y_{t-1} + \alpha_j \cdot s_\theta(y_{t-1},\sigma_j,x_I),
\end{equation}
where $\alpha_j = \epsilon\cdot \sigma_j^2/\sigma_L^2$ is the step size. 
% Since the score function is the gradient of the log-likelihood estimation, the noise-free Langevin equation (Eq. \eqref{Langevin1}) also represents the process of gradient descent. 
% We call the inference algorithm following Eq. \ref{Langevin1} as \textit{Gradient Refinement}.

Next, we analyze the convergence results of the noise-free Langevin equation.
Assume that we have a well-trained score network that can estimate the gradient accurately. For a prediction pair $(x_I,y_I)$,
the noise-free Langevin equation is,
\begin{equation}
\label{Langevin2}
    y_t=y_{t-1} - r_L \cdot (y_{t-1}-y_I),
\end{equation}
where $r_L = \epsilon / \sigma_L^2$ is called the refinement rate.  
% According to the modified Langevin equation, 
Similar to GBM,
it iteratively reduces the difference between $y_t$ and the target $y_I$, and then lets $y_t$ converge to $y_I$ as $t\rightarrow\infty$. 
% Therefore, the inference process based on the noise-free Langevin equation can be regarded as eliminating the error between the current predicted state $y_t$ and the targeted state $y_i$. 
Simultaneously, the estimation of score network $s_\theta(y,\sigma_j,x_I)$ is the error estimation weighted by $\frac{1}{\sigma_i}$. 
We call the inference algorithm following Eq. \eqref{Langevin2} as \textit{error refinement}.
\begin{technique}
\label{tech:11111}
(\textit{Error Refinement})
In the inference stage, the denoising process follows the GBM framework and works as the noise-free Langevin equation (Eq. \eqref{Langevin1}) without the term in the original Langevin equation (Eq. \eqref{Langevin}).
\end{technique}

From this perspective, adding a set of noises aims to fully train the score network to obtain more accurate estimates. 
In \cite{song2020improved}, the authors proposed a strategy to guide the selection of noise where the score network can be more accurately estimated. 
This noise level is a geometric progression based on the “three sigma rule of thumb” \cite{grafarend2006linear}, which can ensure that the samples from $p_{\sigma_i}(x)$ will cover high density regions of $p_{\sigma_{i-1}}(x)$. Besides, for the initial noise scale, the authors suggest choosing $\sigma_1$ to be as large as the maximum Euclidean distance between all pairs of training data points, i.e., $\sigma_1\geq \max_{y_i,y_j}\Vert y_i-y_j  \Vert_{2}$.
Score models based on geometric progression have achieved good performance in a variety of tasks \cite{yang2023diffusion}. 
Therefore, we also adopt a geometric progression noise level and the technique for choosing the initial noise scale.

% Furthermore, we explore the configuration for different steps targeted at classification and regression tasks.

\subsection{Refinement steps}

% After setting the training target, a reasonable configuration of the denoising process for prediction will affect the inference time and prediction accuracy. 

It has been observed that a notorious shortcoming of NCSN and other score-based generative models is their relatively long inference time \cite{cao2024survey}.
In \cite{song2019generative}, the refinement steps for different levels of noise are usually the same value. 
% However, the last denoising step is the most important 
% as discussed above. 
However, according to the \textit{error refinement} equation (Eq. \eqref{Langevin2}), 
the final stage is crucial when deploying a hierarchical denoising process. 
When the last level of noise can be estimated accurately enough, as the denoising step size increases, we can be confident that the predicted value converges around the ground truth.
On the contrary, even if the previous levels of noise are estimated accurately, convergence cannot be guaranteed if the last level is not accurate. 
Hence,
the same refinement steps may waste inference time on denoising processes with large levels of noise.
Additionally, even though the authors in \cite{song2020improved} analyzed how to select a suitable refinement step for different noise levels, it is based on the different inference process from
% follows Alg. \ref{alg:ald} 
the \textit{error refinement} 
following Eq. \eqref{Langevin1}. Therefore, we propose our strategy for step setting. 

Give an input-target pair $(x_I,y_I)\in \mathcal{D}$, following the design of noise level \cite{song2020improved}, the samples from $p_{\sigma_i}(y_I)$ will cover high density regions of $p_{\sigma_{i-1}}(y_I)$. Therefore, in the process of training, there will be more training samples from regions with low noise levels. Additionally, for a perturbed data $\widetilde{y}_I$, since the score network $s_{\theta} (\widetilde{y}_I, \sigma_i, x_I)$ estimates $(\widetilde{y}_I-y_I) / \sigma_i^2$, the estimation based on small noise level is more accurate, when the loss is small enough. 
Due to the \textit{error refinement} continuously reducing the error between the current prediction and the ground truth, a simple idea is to select noise levels based on the current error.
% choosing noise levels based on errors is more helpful for predictions.
Based on this, when the error is reduced to high density areas of noise level $\sigma_{i+1}$, we believe the score estimated by the noise level $\sigma_{i+1}$ to be more accurate than $\sigma_i$. For the current estimated prediction, we have the following proposition, 
\begin{proposition}
\label{Pro1}
    For an input-target pair $(x_I,y_I) \in \mathcal{D}$, let $y_0$ denote the initial prediction point. After $t$ steps of denoising,
    the current estimated prediction $y_t$ satisfies,
    \begin{equation}
        y_t = (1-r_L)^{t} \cdot (y_0 - y_I) + y_I,
    \end{equation}
    where $r_L = \epsilon / \sigma_L^2$ is the refinement rate.
\end{proposition}
When $\vert 1-r_L \vert<1$ and $t\rightarrow \infty$, we have $(1-r_L)^t \cdot(y_0(\sigma_L)-y_I)\rightarrow 0$, i.e., $\lim_{t\rightarrow\infty}y_t = y_I$. Therefore, to ensure the stability and convergence of the \textit{error refinement}, we need to demand $\vert 1-r_L \vert<1$, i.e., the refinement rate satisfies $0<r_L<2$.

Given noise levels $\{\sigma_i\}_{i=1}^L$ and the initial prediction point $y_0(\sigma_i)$ for each noise level $\sigma_i$, let the $t$-th estimated error be $\Vert y_{t} - y_I \Vert_{\infty}$, i.e., the maximum value of the error in all dimensions.
Based on this, we aim to define an end-signal $\beta(\sigma_i)$. When the error estimated by noise level $\sigma_i$ less than this signal $\beta(\sigma_i)$, we jump into using the next noise level to denoise from the current estimated prediction. We call the jump step as the switch time denoted as $t_{\sigma_i}$.
Therefore, we have the following,
\begin{proposition}
\label{Pro2}
    Given a trained score network $s_\theta(y_t,\sigma_i,x_I)$, the end-signal $\beta(\sigma_i)\in \mathbb{R}$ satisfies,
    \begin{equation}
    \label{eq:pro1}
    \begin{split}
            \sigma_i^2 \cdot \Vert s_\theta(y_{t_{\sigma_i}},\sigma_i,x_I) \Vert_{\infty} &< \beta(\sigma_i) \leq \\
            &~~~~~~~~~~\sigma_i^2 \cdot \Vert s_\theta(y_{t_{\sigma_i}-1},\sigma_i,x_I) \Vert_{\infty},
    \end{split}
    \end{equation}
    where $i< L$.
    Moreover, when $1<i<L$, the switch time $t_{\sigma_i}$ satisfies,
    \begin{equation}
    \label{eq:pro2}
        t_{\sigma_i} \leq  \log_{\vert 1-r_L \vert}(\frac{\beta(\sigma_i)}{\beta(\sigma_{i-1})}).
    \end{equation}
    Specially, when $i=1$, the switch time $t_{\sigma_i}$ satisfies,
    \begin{equation}
        t_{\sigma_1} \leq  \log_{\vert 1-r_L \vert}(\frac{\beta(\sigma_1)}{\sigma_1}).
    \end{equation}
\end{proposition}

According to Proposition \ref{Pro2}, via comparing the value between the current output of the score network $s_\theta(y_t,\sigma_i,x_I)$ and end-signal $\beta(\sigma_i)$, we can then determine when to switch to the next level of noise for denoising. Additionally, Proposition \ref{Pro2} provides an upper bound estimation of switch time $t_{\sigma_i}$. This upper bound estimation is relevant to refinement rate $r_L$ and end-signal $\beta(\sigma_i)$. Therefore, via selecting a suitable refinement rate and end-signal, we can effectively limit the number of denoising steps on the specified noise level to shorten the time cost.

For example, the authors chose a geometric sequence with $\sigma_1 = 1$ and $\sigma_{10} = 0.01$, denoising step $T=100$, and $\epsilon=2\times10^{-5}$ for image
generation in \cite{song2019generative}. If we set $\beta(\sigma_i) = \gamma\cdot \sigma_i$, based
on Eq. \eqref{eq:pro2}, we can approximately compute that $t_{\sigma_i}\approx3 \ll 100 = T$. Even though the estimation does not consider errors caused by network training, it is still 
significantly less than the given number of denoising steps which can save a lot of inference time. 
In summary, we propose that
\begin{technique}
\label{tech:end-signal}
(End-signal)
By setting an end-signal $\beta(\sigma_i)$, the denoising process ends based on the estimated error rather than a fixed step size. Additionally, based on the selected refinement rate $r_L$ and noise scales $\{\sigma_i\}_{i=1}^L$, the upper bound of switch time $t_{\sigma_i}$ can be estimated when $i<L$.
\end{technique}

% Based on this, we hope to define an end-signal $\beta(\sigma_i)$. When the error estimated by noise level $\sigma_i$ less than this signal $\beta(\sigma_i)$, we jump into using the next noise level to denoise. 

% Therefore, the definition of the end-signal $\beta(\sigma_i)$ is as the following,
% \begin{proposition}
%     Given noise levels $\{\sigma_i\}_{i=1}^L$ and the initial prediction point $y_0(\sigma_i)$ for each noise level $\sigma_i$, let the $t$-th estimated error as $\Vert y_{t} - y_I \Vert_{2}^2$. The end-signal $\beta(\sigma_i)$ is a value when estimated error less than it, then use the next noise level to denoise from current estimated prediction.
%     \begin{equation}
%         \Vert y_{t_{\sigma_i}} - y_I \Vert_{2}^2
%     \end{equation}
%     the end-signal $\beta(\sigma_i)$ satifies.
% \end{proposition}

% Following the design of noise level, when entering a certain range, we believe that the next level of noise estimation is more accurate.

% \todo{add a ref}, $\beta(\sigma_i)$.

% \todo{error bound}
% \begin{proposition}
%     Consider an input-target pair $(x_t,y_t)$, the refinement steps satisfies,
% \end{proposition}

% to find a suitble refinement steps, following the above result. We hope define an end-signal $\beta(\sigma_i)$. When the score less than this signal, we jump into using next noise level to denoise. 

% \begin{proposition}
%     the end-signal $\beta(\sigma)$ satifies.
% \end{proposition}

% \todo{analyze and why it can reduce time.}

\subsection{Refinement rate}

As discussed above, the final prediction result depends directly on the last denoising process. It is determined by three parameters: 1) the last noise level $\sigma_L$; 2) step size parameter $\epsilon$; 3) the number of the last refinement step. The first two parameters are also the components of the refinement rate, which affects the number of refinement steps of previous noise levels according to Proposition \ref{Pro2}. Therefore, we need to specify the relationship between these parameters and the prediction error.

\begin{proposition}
\label{pro3}
    Let $e(y_t, x_I)$ represent the error between the estimated score and the ground truth score, i.e.,
\begin{equation*}
     e(y_t, x_I) = s_\theta(y_t,\sigma_L,x_I) + \frac{y_t-y_{I}}{\sigma_L^2}.
\end{equation*}
After $t$ times of denoising, the current estimated prediction $y_t$ satisfies
\begin{equation}
\label{pro3:eq1}
    y_t =  (1-r_L)^t \cdot (y_{0}(\sigma_L) - y_I) + y_I + \sum_{i = 0}^{t-1}(1-r_L)^i \cdot E_{i}(x_I) ,
\end{equation}
where $E_{i}(x_I)= \alpha_L\cdot e(y_{i},x_I)$ represents the neural network error.
\end{proposition}
To analyze the constraints of these parameters, we need the value $y_t$ in Eq. \eqref{pro3:eq1} to converge as much as possible to ground truth $y_I$. However, when $t\rightarrow\infty$, $y_t$ cannot converge to $y_I$ due to the estimated error $E_i(\cdot)$. 
Therefore, considering the prediction error between current $y_t$ and $y_I$,
we have the following proposition.
\begin{proposition}
\label{pro4}
Assume the last initial denoising point $y_0(\sigma_L)$ follows Technique \ref{tech:end-signal}, i.e., $\Vert y_0(\sigma_L) -y_I \Vert_{\infty} < \beta(\sigma_{L-1})$. Let $E\in \mathbb{R}$ be the upper bound of $E_i(x_I)$ for all $i\geq 0$ and $(x_I,y_I) \in \mathcal{D}$, i.e., $\Vert E_i(x_I)\Vert _2< E$. For an end-signal $\beta(\sigma_L)$, when 
\begin{equation}
\label{pro4:eq1}
\vert 1-r_L \vert^t \cdot \sqrt{d}\cdot \beta(\sigma_{L-1}) + E \cdot  \frac{1-\vert1-r_L\vert^t}{r_L} < \beta(\sigma_L),
\end{equation}
where $d$ is the number of dimensions of $y_I$.
we can guarantee that $\Vert y_t - y_I \Vert _2 < \beta(\sigma_L) $.
\end{proposition}
We can use the loss value obtained during training to approximate the error $E$ caused by the network.
The end-signal $\beta(\sigma_L)$ can also be regarded as the error bound we hope to achieve iteratively. The end-signal can be set based on task accuracy requirements. After the end-signal is selected, we can choose all three parameters based on the condition in Eq. \eqref{pro4:eq1}. Additionally, the first term $\vert 1-r_L \vert^t \cdot \sqrt{d}\cdot \beta(\sigma_{L-1})$ decreases as $t$ increases. On the contrary, the second term $E \cdot  \frac{1-\vert1-r_L\vert^t}{r_L}$ increases as $t$ increases. On the other hand, we can choose a suitable $T$ where the second term has a higher order than the first term, i.e., $(\vert 1-r_L \vert^T \cdot \sqrt{d}\cdot \beta(\sigma_{L-1}))/ (E \cdot  \frac{1-\vert1-r_L\vert^T}{r_L}) \ll 1$. When inference time $t\geq T$, we can ensure that the error of the current prediction result is dominated by the network training error.
This approach can balance the error caused by the neural network and the error caused by iteration as much as possible. In this case, $t\geq \log_{\vert 1-r_L \vert}\frac{E}{\sqrt{d}\cdot\beta(\sigma_{L-1})\cdot r_L + E}$.
In summary, 
\begin{technique}
\label{tech:three}
(Refinement rate)
By setting an end-signal $\beta(\sigma_L)$, choose suitable $r_L$ and $T$ in the sampling process following Eq. \eqref{pro4:eq1}. Additionally, these parameters can make the second term have a higher order of magnitude than the first term in Eq. \eqref{pro4:eq1}.
\end{technique}

\subsection{Weight of Loss}

The choice of coefficient $\lambda(\cdot)$ in Eq \eqref{myloss} can affect how well the network is trained for different noise levels. In \cite{song2019generative}, the authors aimed to let the value of $\lambda(\sigma_i)l(\theta;\sigma_i)$ in Eq \eqref{NCSNloss2} at the same order of magnitude for all $\{\sigma_i\}_{i=1}^L$. To achieve this, they chose $\lambda(\sigma_i)$ as $\sigma_i^2$. Setting the same magnitude, the loss of the score network learns to a similar degree for each level of noise in each back-propagation and gradient descent. 

% However, for \textit{error refinement} algorithm, does the magnitude of noise at each level need to be the same?

% According to the \textit{error refinement} equation (Eq. \eqref{Langevin2}), 
% the final stage is crucial, when deploying hierarchical denoising process. 
% When the last level of noise can be estimated accurately enough, as the denoising step size increases, we can be confident that the predicted value converges around the ground truth.
% On the contrary, even if the previous levels of noise are estimated accurately, convergence cannot be guaranteed if the last level is not on time. 

When applying Techniques \ref{tech:end-signal} and \ref{tech:three}, the number of denoising steps for different noises is different, and the last step of denoising is the most important.
Based on this, to obtain a better estimation of the score function for the last noise level, an intuitive idea is to enhance the training weight in the loss function \cite{caruana1997multitask}.
Therefore, we choose a different coefficient $\lambda(\sigma_i)$ in Eq \eqref{myloss}, specifically taken as $\sigma_i^k$ and $k<2$. Since $\{\sigma_i\}_{i=1}^L$ is a geometric progression and $\sigma_i^2 l(\theta;\sigma_i)\propto 1$, $\sigma_i^k l(\theta;\sigma_i)\propto \sigma_i^{k-2}$ is also a geometric progression getting the maximum value when $i=L$, i.e., the last noise level. Based on this, we propose a coefficient selection strategy.
\begin{technique}
\label{tech:lossssss}
(Weigh of Loss)
Choosing suitable coefficients $\lambda(\sigma_i)$ in the loss (Eq. \eqref{myloss}) increases the magnitude of the value $\lambda(\sigma_i)l(\theta; \sigma_i, x,y)$ as $i$ increases, such as $\lambda(\sigma_i)=\sigma_i$.
\end{technique}
By employing Techniques \ref{tech:11111}–\ref{tech:lossssss} together, the complete training and inference procedure are shown as Alg. \ref{alg:1} and Alg. \ref{alg:2}.

\begin{algorithm}[tb]
        \caption{Training}
        \label{alg:1}
        \begin{algorithmic}[1]\footnotesize
            \REQUIRE $\{\sigma_i\}_{i=1}^L$, training set $\mathcal{D}$, loss coefficient $\lambda(\sigma_i)$
            \REPEAT
                \STATE choose $(x,y) \in \mathcal{D}$, $\sigma\in \{\sigma_i\}_{i=1}^L$, and $\widetilde{y}\sim\mathcal{N}(y,\sigma)$
                % \STATE $\widetilde{y}\sim\mathcal{N}(y,\sigma)$
                \STATE Take gradient descent step on
                \STATE $\nabla_\theta \frac{1}{2}\lambda(\sigma_i)\Vert s_{\theta} (\widetilde{y}, \sigma, x) + \frac{\widetilde{y}-y}{\sigma^2} \Vert_2^2$
            \UNTIL \STATE converged
        \end{algorithmic}
    \end{algorithm}

\begin{algorithm}[tb]
        \caption{Inference}
        \label{alg:2}
        \begin{algorithmic}[1]\footnotesize
            \REQUIRE $\{\sigma_i\}_{i=1}^L,\varepsilon,T, \{\beta(\sigma_i)\}_{i=1}^{L-1}, x_I$
            \STATE Initialize $y_0$
            \FOR{$i \leftarrow 1$ \textbf{to} $L-1$}
                \STATE $\alpha_i \leftarrow \epsilon\cdot\frac{\sigma_i^2}{\sigma_L^2}$
                \REPEAT
                    \STATE $y_t \leftarrow y_{t-1} + \alpha_i s_{\theta}(y_{t-1},\sigma_i, x_I)$
                \UNTIL $\sigma_i^2 \cdot \Vert s_\theta(y_{t},\sigma_i,x_I) \Vert_{\infty} < \beta(\sigma_i)$
                \STATE $y_0 \leftarrow y_T$
            \ENDFOR
            \FOR{$t \leftarrow 1$ \textbf{to} $T$}
                \STATE $y_t \leftarrow y_{t-1} + \epsilon s_{\theta}(y_{t-1},\sigma_L, x_I)$
            \ENDFOR
            \STATE \textbf{return} $y_T$
        \end{algorithmic}
    \end{algorithm}

\vspace{-0.5cm}
\section{Experiment}

\subsection{Toy examples}

\begin{figure*}[h]
    \centering
    \includegraphics[width= 1.0\linewidth]{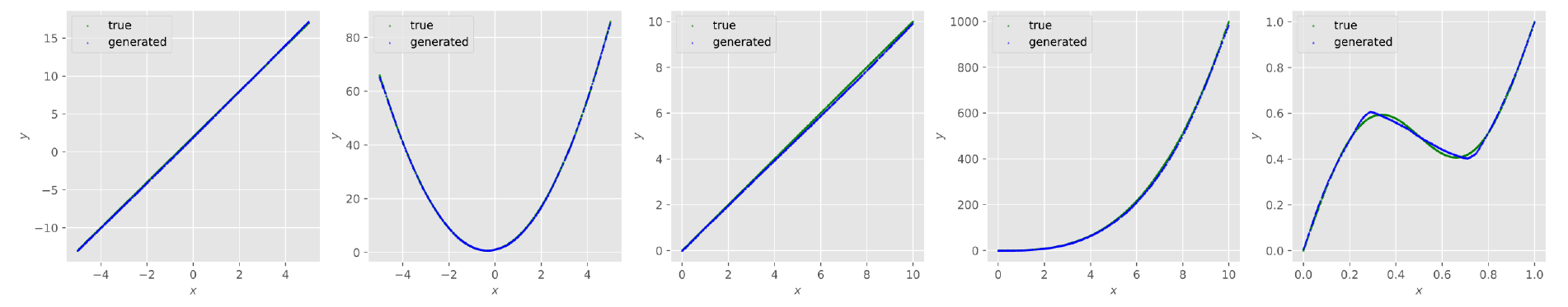}
    \caption{The scatter plots for toy examples. From left to right: linear regression, quadratic regression, log-log linear regression, log-log cubic regression, and sinusoidal regression. The green and blue points represent the true values and the prediction results generated by 1000 samples, respectively.}
    \label{fig:toy}
\end{figure*}

To demonstrate the effectiveness of SSM and all improved techniques, we first perform experiments on 5 selected toy examples (linear regression, quadratic regression, log-log linear regression, log-log cubic regression, and sinusoidal regression) proposed in \cite{han2022card}. We add unbiased normal distribution noise to the original models to obtain noisy data for training, increasing model complexity. We aim to predict accurate regression results for pure input data in the inference process. To evaluate the impact of various technologies, we have designed an ablation experiment that considers three variables: the use of original Langevin dynamics, the implementation of the fast sampling technique as described in Technique \ref{tech:end-signal}, and the application of different coefficients $\lambda(\sigma_i)$ in the loss function, where L1 and L2 represent the degree of noise levels of 1 and 2, respectively. 
The experiment results with the Root Mean Squared Error (RMSE) and inference time about different model settings are shown in Table \ref{tab:ablation_1} and \ref{tab:ablation_2}, respectively. The scatter plots for 5 examples of both true and generated data following the first setting, i.e., L1, w/o noise fast, are shown in Figure \ref{fig:toy}.
We observe that while perturbed data is used, SSM can still fit the regression equation of the original data. The proposed fast sampling Technique \ref{tech:end-signal} significantly reduces the inference time for all tasks. Additionally, since the fast sampling technique ensures the error bound of the current estimated state, it can even achieve the best performance on tasks except for log-log linear regression. Using the noise-free Langevin equation, i.e., w/o noise, performs better on most tasks, indicating that the selection of different coefficients also affects the performance of the model. 
% The selection of different coefficients will also affect the results of the model. 
% A more detailed description of the toy examples, including additional analyses, is presented in the Appendix \ref{ex:toy}. 

% performance metrics

% device 

% 1. Prediction Interval Coverage Probability (PICP); 2. Quantile Interval Coverage Error (QICE).

% inference time.

% use ddim to achieve same time cost 

\subsection{Regression}

We further evaluate our model on 10 UCI regression tasks \cite{dua2017uci}.
We employ the same experimental settings as those used in the CARD model \cite{han2022card}.
% , whose details are provided in the Appendix \ref{ex:UCI}. 
We compare our model with NGboost \cite{duan2020ngboost}, CARD \cite{han2022card}, DBT \cite{han2024diffusion}, non-probabilistic GBM models, including CatBoost \cite{prokhorenkova2018catboost} and XGBoost \cite{chen2016xgboost}, 
% and 
% Bayesian neural network frameworks considered in \cite{han2022card}, including MC Dropout \cite{gal2016dropout}, Deep Ensembles \cite{lakshminarayanan2017simple}, 
and GCDS \cite{zhou2023deep}. 
% Additionally, we further compare our model with non-probabilistic GBM models, including CatBoost \cite{prokhorenkova2018catboost} and XGBoost \cite{chen2016xgboost}.
The experiment results with RMSE are shown in Table \ref{tab:rmse_uci}. 
It is noticed that SSM outperforms existing approaches in RMSE. SSM (L2) obtains the best results in 8 out of 10 datasets. Although the RMSE of SSM (L1) is larger, it has a smaller standard deviation, which means the network is more stable.  
In addition, probabilistic models such as CARD predict the probability of distribution, while SSM predicts the points directly. This also leads to a lower standard deviation of SSM on most tasks.  

% A deeper discussion is provided in the Appendix \ref{ex:com:card}.

\subsection{Classification}

For classification tasks, we compare our model with CARD on CIFAR-10 and CIFAR-100, focusing on both accuracy and the inference time \cite{krizhevsky2009learning}. 
% Similar to CARD, our motivation is to demonstrate that the improved techniques we propose can better adapt score-based generative models to supervised learning rather than achieve state-of-the-art performance. 
% The experimental setup is shown in Appendix \ref{ex:re:cifar}. 
Table \ref{tab:cifar10_comparison} shows the comparison results of accuracy for CIFAR-10 and CIFAR-100 classification.
We observe that SSM can achieve excellent performance for prediction accuracy with lower variance.
This shows that our method has more stable prediction results when selecting different random number seeds.
% Additionally, compared with CARD, 
% SSM can significantly shorten the inference time while ensuring performance, which is shown in Table \ref{tab:cifar10-time}. 
Additionally, SSM can significantly shorten the inference time while ensuring performance compared with CARD, which has the same network structure. 

% More performance comparisons about accuracy and inference time for both CIFAR-10 and CIFAR-100 are provided in Appendix \ref{ex:re:cifar}.

% \begin{table}[h]
%     \centering
%     \caption{Comparison of accuracy (in \%) for CIFAR-10 classification.}
%     \resizebox{\textwidth}{!}{
%     \begin{tabular}{c|c|c|c|c|c|c|c|c|c}
%         \toprule
%         Model & CMV-MF-VI & CM-MF-VI & CV-MF-VI & MF-VI & MC Dropout & MAP & CARD & SSM (L1) & SSM (L2) \\
%         \hline
%         Accuracy & 86.25 $\pm$ 0.06 & 86.66 $\pm$ 0.24 & 79.78 $\pm$ 0.30 & 77.08 $\pm$ 1.14 & 83.64 $\pm$ 0.28 & 84.69 $\pm$ 0.35 & 90.93 $\pm$ 0.02 & 90.91 $\pm$ 0.00 & \textbf{90.99 $\pm$ 0.00}
%  \\
%         \bottomrule
%     \end{tabular}
%     }
%     \label{tab:cifar10_comparison}
% \end{table}

% \begin{table}[h]\footnotesize
%     \centering
%     \caption{Comparison of mean inference time (in second (s)) for one batch on CIFAR-10 dataset.}
%     \begin{tabular}{c|c|c|c}
%         \toprule
%         Model & CARD & SSM (L1) & SSM (L2) \\
%         \hline
%         Time & \textbf{91.79 $\pm$ 0.09} & 19.63 $\pm$ 1.21 & \textbf{15.43 $\pm$ 1.12}\\
%         \bottomrule
%     \end{tabular}
%     \label{tab:cifar10-time}
% \end{table}

% \todo{The main text of a submitted paper is limited to nine content pages, including all figures and tables. Additional pages containing references don’t count as content pages. If your submission is accepted, you will be allowed an additional content page for the camera-ready version.}
\begin{table}[h]
    \centering
    \setlength{\tabcolsep}{1mm}\footnotesize
    \begin{tabular}{c|c|c|c|c}
    \hline
        \multirow{1}{*}{Model} & ~& \multicolumn{1}{c|}{CARD} & \multicolumn{1}{c|}{SSM (L1)} &  \multicolumn{1}{c}{SSM (L2)}\\
        \hline
        %&  & Accuracy & PAvPU & Accuracy & PAvPU & Accuracy & PAvPU\\ 
        CIFAR &  Accuracy &  90.93 $\pm$ 0.02 & 90.91 $\pm$ 0.00 & \textbf{90.99 $\pm$ 0.00}\\
        -10&Time & 50.30 $\pm$ 0.30 & 19.63 $\pm$ 1.21 & \textbf{15.43 $\pm$ 1.12}\\\hline
        CIFAR &Accuracy & 71.42 $\pm$ 0.01 & \textbf{71.51 $\pm$ 0.00} &  71.38 $\pm$ 0.00 \\
        -100&Time & 90.58 $\pm$ 0.12 & 32.30 $\pm$ 0.60 & \textbf{27.16 $\pm$ 0.19}\\
        \hline
    \end{tabular}
    \caption{
    Accuracy (in \%) and inference time (in second (s)) on CIFAR-10 and CIFAR-100.}
    \label{tab:cifar10_comparison}
\end{table}

\begin{table*}[h]
    \centering
    \setlength{\tabcolsep}{1mm}\footnotesize
    % \resizebox{\textwidth}{!}{
    \begin{tabular}{c|cccc|cccc}
    \hline
        \multirow{3}{*}{\diagbox[width=3cm]{Task}{Technique}} &  \multicolumn{8}{c}{RMSE $\downarrow$} \\ \cline{2-9}
        & \multicolumn{4}{c|}{$\lambda(\sigma_i) = \sigma_i$ (L1)} & \multicolumn{4}{c}{$\lambda(\sigma_i) = \sigma_i^2$ (L2)} \\\cline{2-9}
        % &  \checkmark,\checkmark,\checkmark & \texttimes, \checkmark, \checkmark & \texttimes, \checkmark, \texttimes &\checkmark, \checkmark, \texttimes & \texttimes, \texttimes, \checkmark  & \checkmark, \texttimes, \texttimes & \checkmark, \texttimes, \checkmark & \texttimes, \texttimes, \texttimes \\\hline
        & w/o noise, fast & w/ noise, fast & w/ noise &  w/o noise & w/ noise, fast & w/o noise & w/o noise, fast & w/ noise \\\cline{1-1}
        Linear & $\textbf{0.0667}$ & $0.0668$ &  $0.1767$ & $0.0773$ & $0.0685$ & $0.0822$  & $0.0684$ & $0.1764$  \\\cline{1-1}
        Quadratic & $0.1197$ & $0.1198$ &  $0.1103$ & $0.1296$ & $0.0951$ & $0.1026$  & $\textbf{0.0950}$ & $0.1139$ \\\cline{1-1}
        Log-Log Linear & $0.0783$ & $0.0781$ & $0.0857$  & $\textbf{0.0654}$ & $0.0793$ & $0.0885$  & $0.0794$ & $0.0699$  \\\cline{1-1}
        Log-Log Cubic & $3.2704$ & $3.2213$ &  $3.3903$ & $3.9410$ & $\textbf{1.9919}$ & $2.9880$  & $2.0050$ & $3.2718$  \\\cline{1-1}
        Sinusoidal & $0.0096$  & $0.0097$ &  $ 0.0098$ & $0.0094$ & $0.0035$ & $0.0033$  & $\textbf{0.0033}$ & $0.0034$  \\
        \hline
    \end{tabular}
    % }
    \caption{RMSE on different ablation settings on toy examples. W/ and w/o noise represent whether the noise term is included in Langvin dynamics; fast denotes the use of Technique \ref{tech:end-signal};
    L1 and L2 represent the degree of noise levels of 1 and 2, respectively.}
    \label{tab:ablation_1}
\end{table*}

\begin{table*}[h]
    \centering
    \setlength{\tabcolsep}{1mm}\footnotesize
    % \resizebox{\textwidth}{!}{
    \begin{tabular}{c|cccc|cccc}
        \hline
        \multirow{3}{*}{\diagbox[width=3cm]{Task}{Technique}} &  \multicolumn{8}{c}{Inference Time (s) $\downarrow$} \\ \cline{2-9}
        & \multicolumn{4}{c|}{$\lambda(\sigma_i) = \sigma_i$ (L1)} & \multicolumn{4}{c}{$\lambda(\sigma_i) = \sigma_i^2$ (L2)} \\\cline{2-9}
        % &  \checkmark,\checkmark,\checkmark & \texttimes, \checkmark, \checkmark & \texttimes, \checkmark, \texttimes &\checkmark, \checkmark, \texttimes & \texttimes, \texttimes, \checkmark  & \checkmark, \texttimes, \texttimes & \checkmark, \texttimes, \checkmark & \texttimes, \texttimes, \texttimes \\\hline
        & w/o noise, fast & w/ noise, fast & w/ noise &  w/o noise & w/ noise, fast & w/o noise & w/o noise, fast & w/ noise \\\cline{1-1}
        Linear & $0.0385$ & $\textbf{0.0354}$ &  $0.1404$ & $0.1114$ & $0.0376$ & $0.1106$  & $0.0366$ & $0.1404$  \\\cline{1-1}
        Quadratic & $\textbf{0.2743}$ & $0.2774$ &  $1.0957$ & $0.9622$ & $0.3262$ & $0.9631$  & $0.3252$ & $1.0909$  \\\cline{1-1}
        Log-Log Linear & $0.4151$ & $0.4176$ & $1.1151$ & $0.9771$ & $0.3425$ & $0.9647$  & $\textbf{0.3415}$ & $1.0930$  \\\cline{1-1}
        Log-Log Cubic & $0.1311$ & $0.1319$ &  $0.5741$ & $0.5015$ & $\textbf{0.0987}$ & $0.4996$  & $0.0998$ & $0.5843$  \\\cline{1-1}
        Sinusoidal & $0.1774$ & $\textbf{0.1766}$ &  $1.0863$ & $0.9545$ & $0.2103$ & $0.9590$  & $0.2066$ & $1.1023$  \\
        \hline
    \end{tabular}
    % }
    \caption{Inference time on different ablation settings on toy examples. W/ and w/o noise represent whether the noise term is included in Langevin dynamics; fast denotes the use of Technique \ref{tech:end-signal}; L1 and L2 represent the degree of noise levels of 1 and 2, respectively.}
    \label{tab:ablation_2}
\end{table*}

\begin{table*}[h]
    \centering
    \setlength{\tabcolsep}{1mm}\footnotesize
    % \resizebox{\textwidth}{!}{
    \begin{tabular}{c|c|c|c|c|c|c|c|c}
    \hline
        Dataset & \multicolumn{8}{c}{RMSE $\downarrow$} \\
        & NGboost & CatBoost & XGBoost & GCDS & CARD & DBT & SSM (L1) & SSM (L2) \\
        \hline
        Boston & 2.94 $\pm$ 0.53 & 2.19 $\pm$ 0.09 & 2.23 $\pm$ 0.04 & 2.75 $\pm$ 0.58 & 2.61 $\pm$ 0.63  & 2.73 $\pm$ 0.62 & 2.06 $\pm$ 0.19 & \textbf{2.04 $\pm$ 0.14}\\
        
        Concrete & 5.06 $\pm$ 0.61 & 4.75 $\pm$ 0.16 & 4.62 $\pm$ 0.13 & 5.39 $\pm$ 0.55 & 4.77 $\pm$ 0.46 & 4.56 $\pm$ 0.50 &5.88 $\pm$ 1.00 & \textbf{3.20 $\pm$ 1.00} \\
       
        Energy & 0.46 $\pm$ 0.06 & 0.32 $\pm$ 0.02 & \textbf{0.23 $\pm$ 0.01} & 0.64 $\pm$ 0.09 & 0.52 $\pm$ 0.07  & 0.52 $\pm$ 0.07 & 0.36 $\pm$ 0.09 & 0.31 $\pm$ 0.02\\
        
        Kin8nm & 16.03 $\pm$ 0.04 & 9.31 $\pm$ 0.15 & 11.89 $\pm$ 0.17 & 8.88 $\pm$ 0.42 & 6.32 $\pm$ 0.18 & 7.04 $\pm$ 0.23 &8.30 $\pm$ 0.10 & \textbf{4.66 $\pm$ 0.30}\\
        
        Naval & 0.13 $\pm$ 0.00 & 0.13 $\pm$ 0.00 & 0.10 $\pm$ 0.00 & 0.14 $\pm$ 0.05 & 0.02 $\pm$ 0.00  & 0.07 $\pm$ 0.01 & 0.10 $\pm$ 0.00 & \textbf{0.01 $\pm$ 0.00} \\
        
        Power & 3.79 $\pm$ 0.18 & 3.41 $\pm$ 0.03 & \textbf{3.22 $\pm$ 0.03} & 4.11 $\pm$ 0.16 & 3.93 $\pm$ 0.17 & 3.95 $\pm$ 0.16 & 4.25 $\pm$ 0.01 & 3.67 $\pm$ 0.33 \\
        
        Protein & 4.33 $\pm$ 0.03 & 3.83 $\pm$ 0.02 & \textbf{3.53 $\pm$ 0.01} & 4.50 $\pm$ 0.02 & 3.73 $\pm$ 0.01 & 3.81 $\pm$ 0.04 & 5.20 $\pm$ 0.01 & 3.95 $\pm$ 0.43\\
        
        Wine & 0.63 $\pm$ 0.04 & 0.58 $\pm$ 0.02 & 0.57 $\pm$ 0.01 & 0.66 $\pm$ 0.04 & 0.63 $\pm$ 0.04 & 0.61$\pm$ 0.04 & 0.68 $\pm$ 0.01 & \textbf{0.46 $\pm$ 0.21}\\
        
        Yacht & 0.50 $\pm$ 0.20 & 0.44$\pm$ 0.07 & 0.43 $\pm$ 0.08 & 0.79 $\pm$ 0.26 & 0.65 $\pm$ 0.25  & 1.08 $\pm$ 0.39 & 0.56 $\pm$ 0.16 &\textbf{0.37 $\pm$ 0.21}\\
        
        Year & 8.94 $\pm$ NA & 8.94 $\pm$ NA & 8.77 $\pm$ NA & 9.20 $\pm$ NA & \textbf{8.70 $\pm$ NA} & 8.81 $\pm$ NA & 10.14 $\pm$ NA & 10.04 $\pm$ NA \\
        \hline
        \hline
        \# best & 0 & 0 & 3 & 0&1&0 & 0&6\\
        \hline
    \end{tabular}
    % }
    \caption{RMSE results for UCI regression tasks. For Kin8nm and Naval dataset, the results are multiplied by 100 to match the scale of others.}
    \label{tab:rmse_uci}
\end{table*}

\vspace{-0.5cm}
\section{Conclusion}

In this paper, we proposed Supervised Score-based Model (SSM), score-based gradient boosting model using denoising score matching for supervised learning.
First, we analyzed the connection between SSM and GBM and showed that 
the iterative equation of GBM can be converted to a noise-free Langevin equation.
Then, we provided a theoretical analysis on learning and sampling for SSM to balance inference time and prediction accuracy with noise-free Langevin sampling. Furthermore, we showed the effectiveness of the proposed techniques on selected toy examples. Lastly, we compared SSM with existing GBM models and diffusion models, 
% including NGboost, CARD, DBT, GCDS, CatBoost, and XGBoost, 
for several regression and classification tasks. The experimental results show that our model outperforms in both accuracy and inference time.

\FloatBarrier
\newpage
\clearpage
\section{Acknowledgments}
This research is supported by the National Research Foundation, Singapore, and Infocomm Media Development Authority under its Future Communications Research \& Development Programme, Defence Science Organisation (DSO) National Laboratories under the AI Singapore Programme (FCP-NTU-RG-2022-010 and FCP-ASTAR-TG-2022-003), Singapore Ministry of Education (MOE) Tier 1 (RG87/22), the NTU Centre for Computational Technologies in Finance (NTU-CCTF), and Seitee Pte Ltd, and the National Research Foundation, Prime Minister’s Office, Singapore under its Campus for Research Excellence and Technological Enterprise (CREATE) programme.

%\newpage
\bibliography{aaai25}

% \newpage
% \clearpage
% \input{checklist}

\newpage
\clearpage
\appendix
\setcounter{proposition}{0}
\section{Proofs}
\label{ex:proof}

\setcounter{equation}{11}
\begin{proposition}
    For an input-target pair $(x_I,y_I) \in \mathcal{D}$, let $y_0$ denote the initial prediction point. After $T$ steps of denoising,
    the current estimated prediction $y_T$ satisfies,
    \begin{equation}
         y_t = (1-r_L)^{t} \cdot (y_0 - y_I) + y_I,
    \end{equation}
    where $r_L = \epsilon / \sigma_L^2$ is the refinement rate.
\end{proposition}
\begin{proof}
    According to the \textit{error refinement} equation (Eq. \eqref{Langevin2}), we have
\begin{equation*}
    y_t=y_{t-1} - r_L \cdot (y_{t-1}-y_I),
\end{equation*}
    where $r_L = \epsilon / \sigma_L^2$. Hence,
\begin{align*}
    y_t& =(1-r_L)\cdot y_{t-1} + r_L\cdot y_{I},\\
    \Rightarrow y_t - y_I &= (1-r_L)\cdot y_{t-1} + (r_L-1)\cdot y_I\\
     & = (1-r_L) \cdot (y_{t-1} - y_I).
\end{align*}
Therefore, when the initial prediction point is $y_0$, we have
\begin{align*}
    y_t & =(1-r_L)^{t} \cdot(y_0 - y_I) + y_I.
\end{align*}
\end{proof}

\begin{proposition}
    Given a trained score network $s_\theta(y_t,\sigma_i,x_I)$, the end-signal $\beta(\sigma_i)\in \mathbb{R}$ satisfies,
    \begin{equation}
    \label{eq:pro1}
    \begin{split}
        \sigma_i^2 \cdot \Vert s_\theta(y_{t_{\sigma_i}},\sigma_i,x_I) \Vert_{\infty} &< \beta(\sigma_i) \leq \\
        &~~~~~~~~\sigma_i^2 \cdot \Vert s_\theta(y_{t_{\sigma_i}-1},\sigma_i,x_I) \Vert_{\infty},
    \end{split}
    \end{equation}
    where $i< L$.
    Moreover, when $1<i<L$, the switch time $t_{\sigma_i}$ satisfies,
    \begin{equation}
        t_{\sigma_i} \leq  \log_{1-r_L}(\frac{\beta(\sigma_i)}{\beta(\sigma_{i-1})}).
    \end{equation}
    Specially, when $i=1$, the switch time $t_{\sigma_i}$ satisfies,
    \begin{equation}
        t_{\sigma_1} \leq  \log_{1-r_L}(\frac{\beta(\sigma_1)}{\sigma_1}).
    \end{equation}
\end{proposition}
\begin{proof}
    Since the score network $s_\theta(y_t,\sigma_i,x_I)\approx \frac{y_I-y_t}{\sigma_i^2}$ and the estimated error is $\Vert y_{t} - y_I \Vert_{\infty}$. Therefore, we have
    \begin{align*}
        \Vert y_{t} - y_I \Vert_{\infty} \approx \sigma_i^2 \cdot \Vert s_\theta(y_t,\sigma_i,x_I) \Vert_{\infty}.
    \end{align*}
    Based on the definitions of the end-signal $\beta(\sigma_i)$ and the switch time $t_{\sigma_i}$, we have
    \begin{align*}
        \Vert y_{t_{\sigma_i}} - y_I \Vert_{\infty} &< \beta(\sigma_i) \leq \Vert y_{t_{\sigma_i}-1} - y_I \Vert_{\infty}.
    \end{align*}
    Therefore, we immediately obtain Eq. \eqref{eq:pro1},
    \begin{align*}
        \sigma_i^2 \cdot \Vert s_\theta(y_{t_{\sigma_i}},\sigma_i,x_I) \Vert_{\infty} &< \beta(\sigma_i) \leq \\
        &~~~~~~~~\sigma_i^2 \cdot \Vert s_\theta(y_{t_{\sigma_i}-1},\sigma_i,x_I) \Vert_{\infty}.
    \end{align*}
    On the other hand, we have 
    \begin{align*}
        \Vert y_{t} - y_I \Vert_{\infty} & = \Vert (1-r_L)^t\cdot (y_0(\sigma_i)-y_I) \Vert_{\infty}\\
        & =\vert 1-r_L \vert^t \cdot \Vert y_0(\sigma_i)-y_I \Vert_{\infty}\\
        & = \vert1-r_L\vert^t \cdot C_{\sigma_i},
    \end{align*}
    where $C_{\sigma_i} = \Vert y_0(\sigma_i)-y_I \Vert_{\infty}$ is only relevant to the initial point $y_0(\sigma_i)$ of the current noise level $\sigma_i$ and ground truth $y_I$.
    When $1\leq i<L$, 
    after switching, we denoise from the current point $y_{t_{\sigma_i}}$, i.e., $y_0(\sigma_{i+1}) = y_{t_{\sigma_i}}$, and thus,
    \begin{align*}
        \Vert y_0(\sigma_{i+1}) - y_I \Vert_{\infty} &< \beta(\sigma_i),\\
        \Rightarrow C_{\sigma_{i+1}} &< \beta(\sigma_i).
    \end{align*}
    For $1<i<L$, the the switch time $t_{\sigma_i}$ satisfies,
    \begin{align*}
        \vert1-r_L\vert^{t_{\sigma_i}}\cdot C_{\sigma_i} < \beta(\sigma_i)\leq \vert1-r_L\vert^{t_{\sigma_i}-1}\cdot C_{\sigma_i}.
    \end{align*}
    Let $t_{\sigma_i}^c$ denote the switch time for the upper bound of $C_{\sigma_i}$, i.e.,
    \begin{align*}
        &\vert1-r_L\vert^{t^c_{\sigma_i}}\cdot \beta(\sigma_{i-1}) < \beta(\sigma_i)\leq \vert1-r_L\vert^{t^c_{\sigma_i}-1}\cdot \beta(\sigma_{i-1}), \\
        \Rightarrow &\vert1-r_L\vert^{t^c_{\sigma_i}} < \frac{\beta(\sigma_i)}{\beta(\sigma_{i-1})} \leq\vert1-r_L\vert^{t^c_{\sigma_i}-1}. \\
    \end{align*}
    Due to $\vert 1-r_L \vert < 1$, we equivalently obtains
    \begin{align*}
    \log_{\vert1-r_L\vert}(\frac{\beta(\sigma_i)}{\beta(\sigma_{i-1})}) < t^c_{\sigma_i} \leq \log_{\vert1-r_L\vert}(\frac{\beta(\sigma_i)}{\beta(\sigma_{i-1})}+1).
    \end{align*}
    We thus have 
    \begin{align*}
        \vert1-r_L\vert^{t^c_{\sigma_i}}\cdot C_{\sigma_i} < \frac{\beta(\sigma_i)}{\beta(\sigma_{i-1})}\cdot  C_{\sigma_i} < \beta(\sigma_i).
    \end{align*} 
    Therefore, $t_{\sigma_i}\leq \inf t^c_{\sigma_i}$, i.e.,
    \begin{align*}
        t_{\sigma_i}\leq\log_{\vert1-r_L\vert}(\frac{\beta(\sigma_i)}{\beta(\sigma_{i-1})}).
    \end{align*} 

     When $i=1$, since the selection of $\sigma_1$ satisfies $\sigma_1\geq \max_{y_i,y_j}\Vert y_i-y_j  \Vert_{2}$, the $\sigma_1$ represents the maximum distance between any two target points in the training set. Therefore, when we choose an initial point $y_0(\sigma_1)$ in sampling space, we can assume the Euclidean distance between the chosen initial point $y_0(\sigma_1)$ and the ground truth $y_I$ is also less than $\sigma_1$, i.e., 
     \begin{align*}
         \Vert y_0(\sigma_1)-y_I  \Vert_{2} < \sigma_1,\\
     \end{align*}
     Therefore, 
     \begin{align*}
         C_{\sigma_{1}} &= \Vert y_0(\sigma_1)-y_I  \Vert_{\infty} \leq \Vert y_0(\sigma_1)-y_I  \Vert_{2} < \sigma_1.
     \end{align*}
     Similarly to the proof above when $1<i<L$, we can also obtain
     \begin{align*}
         t_{\sigma_1} \leq  \log_{\vert1-r_L\vert}(\frac{\beta(\sigma_1)}{\sigma_1}).
     \end{align*}
\end{proof}

\begin{proposition}
    Let $e(y_t, x_I)$ represent the error between the estimated score and the ground truth score, i.e.,
\begin{equation*}
     e(y_t, x_I) = s_\theta(y_t,\sigma_L,x_I) + \frac{y_t-y_{I}}{\sigma_L^2}.
\end{equation*}
After $t$ times of denoising, the current estimated prediction $y_t$ satisfies
\begin{equation}
    y_t =  (1-r_L)^t \cdot (y_{0}(\sigma_L) - y_I) + y_I + \sum_{i = 0}^{t-1}(1-r_L)^i \cdot E_{i}(x_I) ,
\end{equation}
where $E_{i}(x_I)= \alpha_L\cdot e(y_{i},x_I)$ represents the neural network error.
\end{proposition}
\begin{proof}
    According to the noise-free Langevin equation (Eq. \eqref{Langevin1}), we have
\begin{equation*}
    y_t = y_{t-1} + \alpha_L \cdot s_\theta(y_{t-1},\sigma_L,x_i),
\end{equation*}
   where $\alpha_L = \epsilon\cdot \sigma_L^2/\sigma_L^2$. Thus, we have
\begin{align*}
    y_t&=y_{t-1} + \alpha_L \cdot (e(y_{t-1},x_I) - \frac{y_{t-1}-y_I}{\sigma_L^2})\\
    &=y_{t-1} - r_L \cdot (y_{t-1}-y_I) + E_{t-1}(x_I),
\end{align*}
where $r_L = \epsilon / \sigma_L^2$, and $E_{t-1}(x_I)= \alpha_L\cdot e(y_{t-1},x_I)$. Thus, 
\begin{align*}
         y_t - y_I & = (1-r_L) \cdot (y_{t-1} - y_I) + E_{t-1}(x_I) \\
         & = (1-r_L)^2 \cdot (y_{t-2} - y_I) + E_{t-1} + \\
         &~~~~~~~~~~~~~~~~~~~~~~~~~~~~~~~~~~~~~~~~~~~~~~(1-r_L)\cdot E_{t-2}(x_I)\\
         &\cdots\\
         & = (1-r_L)^t \cdot (y_{0}(\sigma_L) - y_I) + \sum_{i = 0}^{t-1}(1-r_L)^i \cdot E_{i}(x_I)
    \end{align*}
Therefore,
\begin{equation*}
    y_t =  (1-r_L)^t \cdot (y_{0}(\sigma_L) - y_I) + y_I + \sum_{i = 0}^{t-1}(1-r_L)^i \cdot E_{i}(x_I).
\end{equation*}
\end{proof}

\begin{proposition}
\label{pro4}
Assume the last initial denoising point $y_0(\sigma_L)$ follows Technique \ref{tech:end-signal}, i.e., $\Vert y_0(\sigma_L) -y_I \Vert_{\infty} < \beta(\sigma_{L-1})$. Let $E\in \mathbb{R}$ be the upper bound of $E_i(x_I)$ for all $i\geq 0$ and $(x_I,y_I) \in \mathcal{D}$, i.e., $\Vert E_i(x_I)\Vert _2< E$. For an end-signal $\beta(\sigma_L)$, when 
\begin{equation}
\label{pro4:eq1}
\vert 1-r_L \vert^t \cdot \sqrt{d}\cdot \beta(\sigma_{L-1}) + E \cdot  \frac{1-\vert1-r_L\vert^t}{r_L} < \beta(\sigma_L),
\end{equation}
where $d$ is the number of dimensions of $y_I$.
we can guarantee that $\Vert y_t - y_I \Vert _2 < \beta(\sigma_L) $.
\end{proposition}

\begin{proof}
    According to Proposition \ref{pro3}, we obtain
    \begin{align*}
        &y_t =  (1-r_L)^t \cdot (y_{0}(\sigma_L) - y_I) + y_I + \sum_{i = 0}^{t-1}(1-r_L)^i \cdot E_{i}, \\
        \Rightarrow & \Vert y_t - y_I \Vert _2 = 
        \\& ~~~~~~~~~\Vert (1-r_L)^t \cdot (y_{0}(\sigma_L) - y_I) +  \sum_{i = 0}^{t-1}(1-r_L)^i \cdot E_{i} \Vert _2.
    \end{align*}
    Since $\Vert y_0(\sigma_L) -y_I \Vert_{\infty} < \beta(\sigma_{L-1})$, according to the norm inequality, we have 
    \begin{align*}
        \Vert y_0(\sigma_L) -y_I \Vert_{2} \leq \sqrt{d}\cdot \Vert y_0(\sigma_L) -y_I \Vert_{\infty} < \sqrt{d}\cdot \beta(\sigma_{L-1}).
    \end{align*}
    Thus, we obtain
    \begin{align*}
        &\Vert y_t - y_I \Vert _2 = \\
        &\Vert (1-r_L)^t \cdot (y_{0}(\sigma_L) - y_I) + \sum_{i = 0}^{t-1}(1-r_L)^i \cdot E_{i} \Vert _2 \\ 
        & \leq \Vert (1-r_L)^t \cdot (y_{0}(\sigma_L) - y_I) \Vert _2 +  \Vert \sum_{i = 0}^{t-1}(1-r_L)^i \cdot E_{i} \Vert _2 \\
        & < \vert 1-r_L \vert^t \cdot \sqrt{d}\cdot \beta(\sigma_{L-1}) + E \cdot \Vert \sum_{i = 0}^{t-1}(1-r_L)^i \Vert _2\\
        & = \vert 1-r_L \vert^t \cdot \sqrt{d}\cdot \beta(\sigma_{L-1}) + E \cdot  \frac{1-\vert1-r_L\vert^t}{r_L}
    \end{align*}
    Therefore, when $\vert 1-r_L \vert^t \cdot \sqrt{d}\cdot \beta(\sigma_{L-1}) + E \cdot  \frac{1-\vert1-r_L\vert^t}{r_L} < \beta(\sigma_L)$,
    we have
       \begin{align*}
\Vert y_t - y_I \Vert _2 < \beta(\sigma_L) .
    \end{align*}
    
\end{proof}
\section{Experimental details}
\label{ex:exdetail}

\begin{table*}[h]
    \centering
    \begin{tabular}[t]{@{}p{0.45\linewidth}@{}p{0.1\linewidth}@{}p{0.45\linewidth}@{}}
        \begin{tabular}[t]{c}
            \textbf{(a) Regression network architecture.} \\
            \hline\hline
            \rule{0pt}{4ex}input: $x, y_t, f_{\phi}(x), t$ \\
            \hline
            \rule{0pt}{4ex}$l_1 = \sigma \left( g_{1,a} \left( x \oplus y_t \oplus f_{\phi}(x) \odot g_{1,b}(t) \right) \right)$ \\
            \hline
            \rule{0pt}{4ex}$l_2 = \sigma \left( g_{2,a}(l_1) \odot g_{2,b}(t) \right)$ \\
            \hline
            \rule{0pt}{4ex}$l_3 = \sigma \left( g_{3,a}(l_2) \odot g_{3,b}(t) \right)$ \\
            \hline
            \rule{0pt}{4ex}output: $g_{4}(l_3)$ \\
        \end{tabular}
        &
        \hspace*{0.05\linewidth}
        &
        \begin{tabular}[t]{c}
            \textbf{(b) Classification network architecture.} \\
            \hline\hline
            \rule{0pt}{4ex}input: $x, y_t, f_{\phi}(x), t$ \\
            \hline
            \rule{0pt}{4ex}$l_{1,x} = \sigma \left( \text{BN}(g_{1,x}(x)) \right)$ \\
            \hline
            \rule{0pt}{4ex}$l_{2,x} = \sigma \left( \text{BN}(g_{2,x}(x)) \right)$ \\
            \hline
            \rule{0pt}{4ex}$l_{3,x} = \text{BN}(g_{1,x}(x))$ \\
            \hline
            \rule{0pt}{4ex}$l_{1,y} = \sigma \left( \text{BN}(g_{1,y}(y_t \oplus f_{\phi}(x)) \odot g_{1,b}(t)) \right)$ \\
            \hline
            \rule{0pt}{4ex}$l_1 = l_{3,x} \odot l_{1,y}$ \\
            \hline
            \rule{0pt}{4ex}$l_2 = \sigma \left( \text{BN}(g_{2,a}(l_1) \odot g_{2,b}(t)) \right)$ \\
            \hline
            \rule{0pt}{4ex}$l_3 = \sigma \left( \text{BN}(g_{3,a}(l_2) \odot g_{3,b}(t)) \right)$ \\
            \hline
            \rule{0pt}{4ex}output: $g_{4}(l_3)$ \\
        \end{tabular}
    \end{tabular}
    \caption{The network architecture in regression and classification experiments. $\oplus$ denotes concatenation; $\odot$ represents Hadamard product; $\sigma$ is Softplus non-linearity; $g(\cdot)$ denotes the fully-connected layer; $l$ is the output of hidden layer; $\text{BN}(\cdot)$ represents the batch normalization.}
    \label{tab:nnstructure}
\end{table*}

\subsection{General Experiment Setup Details}

The code of our project is built and modified from the CARD \footnote{https://github.com/XzwHan/CARD} \cite{han2022card}.
All experiments are conducted on a Linux server using an A100 graphics card.

For the score-based model network architecture, we adopt the same network architecture and the same network parameter settings as those used in \cite{han2022card}, since 
our purpose is to conduct comparative experiments with CARD. Table \ref{tab:nnstructure} shows both regression and classification network architectures. Specifically, the out dimension of all three fully-connected layers is 128 in regression tasks.
We perform Hadamard product between each of the output vector with the corresponding timestep embedding, followed by a Softplus non-linearity, before sending the resulting vector to the next fullyconnected layer. 
The last fully-conected layer $g_4$ map the vector to one with a dimension of 1, as the output for prediction.
For CIFAR-10 classifications, an encoder with three fully-connected layers is first used to obtain a representation with 4096 dimensions. Different from the setup for regression tasks, a Hadamard product between image embedding and response embedding is used and all fully-connected layers with 4096 output dimensions will also use a Hadamard product with a timestep embedding.

In this network architecture, the $f_{\phi}(x)$ is a pre-trained network and will not be jointly trained when training the score-based model. Even though our proposed model dose not write $f_{\phi}(x)$ explicitly, $f_{\phi}(x)$ can be regarded as a part of score network $s_\theta$ with fixed parameters. Hence, such a setting will not affect our above-mentioned proofs and can help us better compare the performance of SSM and CARD. In regression tasks, a feed-forward neural network with two hidden layers, each with 100 and 50 hidden units, is used as the pre-train network. We apply a Leaky ReLU non-linearity with a 0.01 negative slope after each hidden layer.
For the training stage, we set the learning rate to 0.001 for all tasks. We use the AMSGrad \cite{reddi2019convergence} for regression tasks and Adam for classification tasks, which are also the same as the settings in CARD. We also apply a batch size of 256 for all toy regression tasks and classification tasks. The batch size settings of all UCI regression tasks are shown in Table \ref{tab:setupucibat}.

\begin{table*}[h]
    \centering
    % \resizebox{\textwidth}{!}{
    \setlength{\tabcolsep}{1mm}
    \begin{tabular}{c|c|c|c|c|c|c|c|c|c|c}
    \hline
        task & Boston & Concrete & Energy & Kin8nm & Naval & Power & Protein & Wine & Yacht & Year \\\hline
        batch size & 32 & 32 & 32 & 64 & 64 & 64 & 256 & 32 &32&256\\
        \hline
    \end{tabular}
    % }
    \caption{Batch size settings of UCI regression tasks}
    \label{tab:setupucibat}
\end{table*}

% \todo{geng xi jie yi xie}

\subsection{Toy examples}
\label{ex:toy}

The system models of 5 toy examples are present in Table \ref{tab:toymodel}. Each dataset is created by sampling 10240 data points from the system models and then randomly split into training and test sets with an $80\%/20\%$ ratio. Our model setup for toy examples is shown in Table \ref{tab:setuptoy}, where noises are a geometric sequence; $L$ denotes the number of noise, i.e., $\{\sigma_i\}_{i=1}^L$; $\sigma_1$ is the beginning noise selected by approximating maximum training set data distance \cite{song2020improved}; $\sigma_L$ is the ending noise; $Step$ represents the number of denoising steps, which can also ensures the termination of the algorithm when applying fast sampling technique; $\epsilon$ is the step size in Langevin dynamics; $\beta(\sigma_i)$ is the coefficient of end-signal introduced in Technique \ref{tech:end-signal}, which can be computed by $\beta\cdot \sigma_i$; $t_f$ ,$t_b$, and $t_l$ represent the denoising steps computed by Techniques \ref{tech:end-signal} and \ref{tech:three} of the starting noise, middle noise, and ending noise when fast sampling is applied, respectively. In the training phase, we first pre-train $f_{\phi}$ for 100 epochs and then use the pre-trained $f_{\phi}$ to train the SSM model for 5000 epoch. In the testing phase, we sample each input 100 times and take the average as the output.

According to Proposition \ref{pro4}, we need the estimation error of the score network to compute $t_l$. In practice, it is impossible to obtain such an error. Therefore, we use the network's loss function on the training set to approximate the error. However, the error obtained in this way is inaccurate, and using the maximal error to estimate is usually too conservative, which makes the obtained $t_l$ usually too small. Therefore, the number of steps that we choose needs to be greater than $t_l$. Additionally, a step size that is too large will amplify the network error based on Proposition \ref{pro4}, which may also lead to inaccurate predictions. 
As RMSE for linear regression and log-log cubic regression in Table \ref{tab:ablation_1}, using fewer denoising steps, i.e., fast sampling, can achieve better performances than the results obtained by more denoising steps. 
Since the original score-based generative model algorithm requires the number of steps to be as large as possible, this will further amplify the impact of network errors on supervised learning. Therefore, the proposed model SSM provides an effective way to alleviate this problem through fast sampling based on error estimation.

\begin{table*}[h]
    \centering
    \begin{tabular}{c|c|c|c}
    \hline
        Regression Task & Data Generating Function & $x$ & $\epsilon$ \\
        \hline
        Linear & $y = 2x + 3 + \epsilon$ & U($-5$, $5$) & $\mathcal{N}(0, 2^2)$ \\

        Quadratic & $y = 3x^2 + 2x + 1 + \epsilon$ & U($-5$, $5$) & $\mathcal{N}(0, 2^2)$ \\
 
        Log-Log Linear & $y = \exp(x) + \epsilon$ & U($0$, $10$) & $\mathcal{N}(0, 0.15^2)$ \\
        
        Log-Log Cubic & $y = \exp(3\log(x)) + \epsilon$ & U($0$, $10$) & $\mathcal{N}(0, 0.15^2)$ \\
        
        Sinusoidal \cite{bishop1994mixture} & $y = x + 0.3\sin(2\pi x) + \epsilon$ & U($0$, $1$) & $\mathcal{N}(0, 0.08^2)$ \\
        \hline
    \end{tabular}
    \caption{System model of regression toy examples.}
    \label{tab:toymodel}
\end{table*}

\begin{table*}[h]
    \centering
    \begin{tabular}{c|c|c|c|c|c|c|c|c|c}
    \hline
        task & L & $\sigma_1$ & $\sigma_L$ & Step & $\epsilon$ & $\beta$ & $t_f$ & $t_b$ & $t_l$ \\\hline
        Linear & 10 & 20 & 0.01 & 30 & $5*10^-5$ & 0.01 & 7& 1 & 1\\
        Quadratic & 10 & 90 & 0.01 & 30 & $5*10^-5$ & 0.01 & 7 & 1 & 1 \\
        Log-Log Linear & 10 & 20 & 0.01 & 30 & $5*10^-5$ & 0.01 & 7 & 1 & 1\\
        Log-Log Cubic & 10 & 5 & 0.01 & 30 & $5*10^-5$ & 0.01 & 7 & 1 & 1 \\
        Sinusoidal & 10 & 2 & 0.01 & 30 & $5*10^-5$ & 0.01 & 7 & 1 & 1\\
        \hline
    \end{tabular}
        \caption{Setup of SSM on toy example}
    \label{tab:setuptoy}
\end{table*}

\subsection{UCI Regression}
\label{ex:UCI}

We follow the UCI dataset setup in \cite{han2022card}. Our model's parameters setting is illustrated in Table \ref{tab:setupUCI}. We apply the learning rate of 0.001 on all datasets. We train 5 models on each task except UCI-Year due to its size and evaluate the RMSE metric compared with NGboost, CARD, DBT, GCDS, CatBoost, and XGBoost. Additionally, we further compare the proposed model with 
other Bayesian neural networks considered in \cite{han2022card}, including MC Dropout \cite{gal2016dropout}, Deep Ensembles \cite{lakshminarayanan2017simple}
. Via comparing the variance of RMSE for all tasks in Tables \ref{tab:rmse_uci} and \ref{tab:rmse_uci2}, SSMs under different coefficient settings both have smaller variances for most of the tasks. This means that the proposed method has excellent stability in both the training and inference stages. Table \ref{tab:numberUCI1} and Table \ref{tab:numberUCI2} show the average number of denoising step per noise level of SSM (L1) and SSM (L2), respectively. From the results, we observe that the number of denoising steps for the noise levels in the second half will reach the upper bound of $30$ without jumping based on the setting of the end-signals for most tasks. In this case, the error between the current state and the target state cannot be bounded to the specified error range, also due to the error caused by network training. This is also one of the limitations in the upper bound of denoising time estimated in this paper, which is not always consistent with real computation.

\begin{table*}[h]
    \centering
    \setlength{\tabcolsep}{1mm}\footnotesize
    % \resizebox{\textwidth}{!}{
    \begin{tabular}{c|c|c|c|c}
    \hline
        Dataset & \multicolumn{4}{c}{RMSE $\downarrow$} \\
        & MC Dropout & Deep Ensembles & SSM (L1) & SSM (L2) \\
        \hline
        Boston & 3.06 $\pm$ 0.96 & 3.17 $\pm$ 1.05 & 2.06 $\pm$ 0.19 & \textbf{2.04 $\pm$ 0.14}\\
        
        Concrete  & 5.09 $\pm$ 0.60 & 4.91 $\pm$ 0.47 & 5.88 $\pm$ 1.00 & \textbf{3.20 $\pm$ 1.00} \\
       
        Energy  & 1.70 $\pm$ 0.22 & 2.02 $\pm$ 0.32 &  0.36 $\pm$ 0.09 & \textbf{0.31 $\pm$ 0.02}\\
        
        Kin8nm & 7.10 $\pm$ 0.26 & 8.65 $\pm$ 0.47 & 8.30 $\pm$ 0.10 & \textbf{4.66 $\pm$ 0.30}\\
        
        Naval  & 0.08 $\pm$ 0.03 & 0.09 $\pm$ 0.01  & 0.10 $\pm$ 0.00 & \textbf{0.01 $\pm$ 0.00} \\
        
        Power  & 4.04 $\pm$ 0.14 & 4.02 $\pm$ 0.15  & 4.25 $\pm$ 0.01 & \textbf{3.67 $\pm$ 0.33} \\
        
        Protein  & 4.16 $\pm$ 0.12 & 4.45 $\pm$ 0.02 & 5.20 $\pm$ 0.01 & \textbf{3.95 $\pm$ 0.43}\\
        
        Wine  & 0.62 $\pm$ 0.04 & 0.63 $\pm$ 0.04  & 0.68 $\pm$ 0.01 & \textbf{0.46 $\pm$ 0.21}\\
        
        Yacht  & 0.84 $\pm$ 0.27 & 1.19 $\pm$ 0.49 & 0.56 $\pm$ 0.16 &\textbf{0.37 $\pm$ 0.21}\\
        
        Year  & \textbf{8.77 $\pm$ NA }& 8.79 $\pm$ NA & 10.14 $\pm$ NA & 10.04 $\pm$ NA \\
        \hline
        \hline
        \# best & 1 & 0 & 0&9\\
        \hline
    \end{tabular}
    % }
    \caption{RMSE results for UCI regression tasks for the proposed model and Bayesian neural network frameworks. For Kin8nm and Naval dataset, the results are multiplied by 100 to match the scale of others.}
    \label{tab:rmse_uci2}
\end{table*}

% To compare the inference time, we show the mean inference time comparison for one batch on UCI-year regression task of SSM and CARD in Table \ref{tab:uci-year-time}. The results demonstrate our model’s significantly fast inference capability without compromising prediction accuracy.

% \begin{table}[h]
%     \centering
%     \caption{Comparison of mean inference time for one batch (in second (s)) on UCI-year regression tasks.}
%     \begin{tabular}{c|c|c|c}
%         \toprule
%         Model & CARD & SSM (L1) & SSM (L2) \\
%         \hline
%         Time & \textbf{91.79 $\pm$ 0.09} & 0.1404 $\pm$ 0.0001 & \\
%         \bottomrule
%     \end{tabular}
%     \label{tab:uci-year-time}
% \end{table}

\begin{table*}[h]
    \centering
    % \resizebox{\textwidth}{!}{
    \setlength{\tabcolsep}{1mm}
    \begin{tabular}{c|c|c|c|c|c|c|c|c|c|c|c}
    \hline
        task & $n_1$ & $n_2$ & $n_3$ & $n_4$ & $n_5$ & $n_6$& $n_7$& $n_8$ & $n_9$ & $n_{10}$ & t\\\hline
        Boston & 8.80 & 5.00 & 6.40& 7.20 & 12.00& 23.80& 30.00 & 30.00& 30.00& 30.00 & 0.1480 \\
        Concrete & 9.00 & 6.10 & 8.10 & 16.70 & 25.80& 29.10& 30.00 & 30.00& 30.00& 30.00 & 0.1790\\
        Energy & 9.40 & 5.00 & 5.50 & 6.40 & 7.20& 10.40& 17.86 & 24.50& 30.00& 30.00&0.0964\\
        Kin8nm & 8.35 & 8.14 & 9.45 & 10.70 & 20.52& 30.00& 30.00 & 30.00& 30.00& 30.00 & 0.2046\\
        Naval & 8.91 & 5.02 & 6.21 & 7.47 & 11.18& 10.04& 12.97 & 22.31& 28.64& 30.00 & 0.1386\\
        Power & 8.97 & 3.72 & 5.34 & 10.25 & 20.03& 29.66& 30.00 & 30.00& 30.00& 30.00& 0.3342\\
        Protein & 8.88 & 6.13 & 8.60 & 22.01 & 28.73& 30.00& 30.00 & 30.00& 30.00& 30.00 & 0.2232\\
        Wine & 7.93 & 5.40 & 6.13 & 10.40 & 21.33& 7.47& 5.60 & 4.40& 3.73& 30.00 & 0.085\\
        Yacht & 9.80 & 6.80 & 6.20 & 6.00 & 5.80& 8.00& 9.20 & 29.00& 30.00& 30.00 & 0.0754\\
        Year & 30.00  & 6.11& 7.43 & 16.93 & 27.08& 30.00& 30.00 & 30.00& 30.00& 30.00 & 0.0766 \\
        \hline
    \end{tabular}
    % }
    \caption{Average number of denoising steps $n_i$ per noise level and mean inference time $t$ for one batch (in second (s)) on UCI datasets by SSM (L1).}
    \label{tab:numberUCI1}
\end{table*}

\begin{table*}[h]
    \centering
    % \resizebox{\textwidth}{!}{
    \setlength{\tabcolsep}{1mm}
    \begin{tabular}{c|c|c|c|c|c|c|c|c|c|c|c}
    \hline
        task & $n_1$ & $n_2$ & $n_3$ & $n_4$ & $n_5$ & $n_6$& $n_7$& $n_8$ & $n_9$ & $n_{10}$ & t\\\hline
        Boston & 8.80 & 5.00 & 6.20 & 6.80 & 13 & 22.4 & 30.00 & 30.00& 30.00& 30.00 & 0.15437 \\
        Concrete & 8.10 & 6.40 & 7.20 & 9.80 & 15.60& 18.70& 29.10 & 30.00& 30.00& 30.00 & 0.1541\\
        Energy & 8.50 & 4.40 & 4.40 & 4.40 & 5.10& 8.60& 18.90 & 22.70& 30.00& 30.00 & 0.0853\\
        Kin8nm & 9.03& 6.17 & 6.60 & 9.13 & 13.28& 17.65& 22.05 & 28.63& 29.91& 30.00 & 0.1426\\
        Naval & 8.12 & 3.91 & 3.76 & 4.86 & 5.17 & 5.37& 6.00 & 10.57& 21.47& 30.00 & 0.0923\\
        Power & 8.33 & 4.93 & 7.06 & 8.83 & 16.95& 29.15& 29.98 & 30.00& 30.00& 30.00 & 0.3341\\
        Protein & 7.65 & 7.13 & 7.79 & 16.16 & 26.92& 29.81& 30.00 & 30.00& 30.00& 30.00 & 0.2163\\
        Wine & 7.22 & 6.78 & 7.44 & 9.11 & 22.67& 11.56& 7.56 & 8.00& 20.78& 30.00 & 0.3074\\
        Yacht & 8.20 & 4.20 & 4.60 & 4.60 & 5.00& 5.20& 6.20 & 15.40& 27.80& 30.00 & 0.0598\\
        Year & 7.01 & 5.02  & 7.43  &19.06 & 26.94 & 29.97 & 30.00 &30.00 &30.00 &30.00  & 0.0766\\
        \hline
    \end{tabular}
    % }
    \caption{Average number of denoising steps $n_i$ per noise level and mean inference time $t$ for one batch (in second (s)) on UCI datasets by SSM (L2).}
    \label{tab:numberUCI2}
\end{table*}

% \todo{select 3 to draw a curve about accuracy and denoising steps}

% The used UCI datasets and their size are illustrated in Table \ref{tab:dataset_size}.

% \begin{table}[h]
%     \centering
%     \caption{Dataset size (\(N\) observations, \(P\) features) of UCI regression tasks.}
%     \resizebox{\textwidth}{!}{
%     \begin{tabular}{c|c|c|c|c|c|c|c|c|c|c}
%         \toprule
%         Dataset & Boston & Concrete & Energy & Kin8nm & Naval & Power & Protein & Wine & Yacht & Year \\
%         \hline
%         (\(N, P\)) & (506, 13) & (1030, 8) & (768, 8) & (8192, 8) & (11, 934, 16) & (9568, 4) & (45,730, 9) & (1599, 11) & (308, 6) & (515, 345, 90) \\
%         \bottomrule
%     \end{tabular}
%     }
%     \label{tab:dataset_size}
% \end{table}

\begin{table*}[h]
    \centering
    \begin{tabular}{c|c|c|c|c|c|c|c|c|c}
    \hline
        task & L & $\sigma_1$ & $\sigma_L$ & Step & $\epsilon$ & $\beta$ & $t_f$ & $t_b$ & $t_l$ \\\hline
        Boston & 10 & 5 & 0.01 & 30 & $5*10^-5$ & 0.01 & 7 & 1 & 1\\
        Concrete & 10 & 5 & 0.01 & 30 & $5*10^-5$ & 0.01 & 7 & 1 & 1 \\
        Energy & 10 & 5 & 0.01 & 30 & $5*10^-5$ & 0.01 & 7 & 1 & 1\\
        Kin8nm & 10 & 5 & 0.01 & 30 & $5*10^-5$ & 0.01 & 7 & 1 & 1 \\
        Naval & 10 & 2 & 0.01 & 30 & $5*10^-5$ & 0.01 & 7 & 1 & 1\\
        Power & 10 & 20 & 0.01 & 30 & $5*10^-5$ & 0.01 & 7 & 1 & 1\\
        Protein & 10 & 20 & 0.01 & 30 & $5*10^-5$ & 0.01 & 7 & 1 & 1\\
        Wine & 10 & 20 & 0.01 & 30 & $5*10^-5$ & 0.01 & 7 & 1 & 1\\
        Yacht & 10 & 20 & 0.01 & 30 & $5*10^-5$ & 0.01 & 7 & 1 & 1\\
        Year & 10 & 20 & 0.01 & 30 & $5*10^-5$ & 0.01 & 7 & 1 & 1\\\hline
    \end{tabular}
    \caption{Setup of SSM on UCI regression tasks}
    \label{tab:setupUCI}
\end{table*}

\subsection{CIFAR-10 and CIFAR-100 classification}
\label{ex:re:cifar}

CIFAR-10 and CIFAR-100 dataset \cite{krizhevsky2009learning} are classic image classification datasets, which are tasks of 10 categories and 100 categories, respectively. We use the same pre-train networks provided by CARD \cite{han2022card}, where ResNet-18 with a test accuracy of $90.39\%$ is used for CIFAR-10 and ResNet-18 with a test accuracy of $71.37\%$ is used for CIFAR-100.
The parameters in SSM are shown in Table \ref{tab:setupcifar}. During the evaluation, we generate 100 prediction results for each input data using 10 different random number seeds for CIFAR-10 and 10 prediction results for CIFAR-100. The comparisons of accuracy are shown in Tables \ref{tab:cifar10_comparison} and \ref{tab:ACCandPA}. SSM can achieve more stable and better accuracy than CARD. For CIFAR-10 classification in Table \ref{tab:cifar10_comparison}, SSM (L2) can achieve $90.99\%$ accuracy and the accuracy of SSM (L1) is only $0.01\%$ lower than CARD. Table \ref{tab:fashionmnist_piw_ttest} illustrates the comparison of the accuracy in each category of CIFAR-10 classification task. While CARD achieves the highest accuracy in more categories than SSM, its lower all accuracy demonstrates that its generalization ability is not as good as SSM, and it cannot have good learning effects for every category.

The comparison results for CIFAR-100 are shown in Table \ref{tab:ACCandPA}. We observer that SSM (L1) achieves the highest accuracy and improves the accuracy of pre-train model $f_{\phi}$ to $71.51\%$, while CARD can only achieve $0.05\%$ improvement. Moreover, even though the standard deviation of CARD prediction accuracy is already very low, the standard deviation is lower and is almost equal to 0. 
The inference time for CIFAR-100 is also shown in Table \ref{tab:ACCandPA}, where SSM can still outperform CARD. It demonstrates that SSM can significantly fast inference capability without compromising prediction accuracy.

On the other hand, DDIM \cite{song2020denoising} is a commonly used accelerated sampling technique for diffusion-based models in the inference process. 
It is noticed that we do not compare our SSM with CARD sampling by DDIM. The reason is that during the derivation process of the CARD model, to ensure the accuracy of prediction, i.e., the center of the prediction distribution is close to the target value, the authors in \cite{han2022card} proposed a sampling relationship that is different from DDPM \cite{song2020denoising}. Therefore, the DDIM technique cannot be used directly to speed up sampling CARD.
% Moreover, a detailed comparison between SSM and CARD is provided in Appendix \ref{ex:com:card}.

\begin{table*}[h]
    \centering
    \begin{tabular}{c|c|c|c|c|c|c|c|c|c}
    \hline
        task & L & $\sigma_1$ & $\sigma_L$ & Step & $\epsilon$ & $\beta$ & $t_f$ & $t_b$ & $t_l$ \\\hline
        CIFAR-10 & 20 & 1 & 0.01 & 200 & $5\times10^{-5}$ & 0.1 & 3 & 1 & 1\\
        CIFAR-100 & 20 & 1 & 0.01 & 200 & $2\times10^{-5}$ & 0.01 & 21 & 2 & 1 \\
        \hline
    \end{tabular}
    \caption{Setup of SSM on CIFAR-10 and CIFAR-100}
    \label{tab:setupcifar}
\end{table*}

% \begin{table}[h]
%     \centering
%     \caption{Comparison of mean inference time for one batch (in second (s)) on CIFAR-100.}
%     \begin{tabular}{c|c|c|c}
%         \toprule
%         Model & CARD & SSM (L1) & SSM (L2)\\
%         \hline
%         CIFAR-100 & ~ & 32.30 $\pm$ 0.60 & 27.16 $\pm$ 0.19\\
%         \bottomrule
%     \end{tabular}
%     \label{tab:cifarTIME}
% \end{table}

\begin{table*}[h]
    \centering
    \begin{tabular}{c|c|c|c}
    \hline
        Model & CARD & SSM (L1) & SSM (L2)\\
        \hline
        All & 90.95\% & 90.91\% & \textbf{90.99\%}\\
        1 & \textbf{91.00\%} & \textbf{91.00\%} & 90.70\%\\
        2 & 96.00\% & \textbf{96.10\%} & \textbf{96.10\%}\\
        3 & 87.30\% & 88.20\% & \textbf{88.30\%}\\
        4 & \textbf{81.90\%} & 81.30\% & 81.40\%\\
        5 & 93.30\%  & 92.40\% & \textbf{94.10\%}\\
        6 & 84.70\% & \textbf{85.50\%} & 84.80\%\\
        7 & 94.20\% & \textbf{94.50\%} & \textbf{94.50\%}\\
        8 & \textbf{92.80\%} & 92.40\% & 92.30\%\\
        9 & \textbf{95.30\%} & 94.90\% & 95.00\%\\
        10 & \textbf{93.00\%} & 92.80\% & 92.70\%\\
        \hline
        \# best & 5 & 4 & 4 \\
        \hline
    \end{tabular}
    \caption{Comparison of the accuracy of SSM and CARD in each category of CIFAR-10 classification task.}
    \label{tab:fashionmnist_piw_ttest}
\end{table*}

\begin{table*}[h]
    \centering
    \begin{tabular}{c|c|c|c|c}
    \hline
        \multirow{1}{*}{Model} & \multirow{1}{*}{Acc. by $f_{\phi}$} &  \multicolumn{1}{c|}{CARD} & \multicolumn{1}{c|}{SSM (L1)} &  \multicolumn{1}{c}{SSM (L2)}\\
        \hline
        %&  & Accuracy & PAvPU & Accuracy & PAvPU & Accuracy & PAvPU\\ 
        Accuracy & 71.37\% & 71.42 $\pm$ 0.01\% & \textbf{71.51 $\pm$ 0.00\%} &  71.38 $\pm$ 0.00\% \\
        Time & - & 90.58 $\pm$ 0.12 & 32.30 $\pm$ 0.60 & \textbf{27.16 $\pm$ 0.19}\\
        \hline
    \end{tabular}
    \caption{
    Accuracy and mean inference time for one batch (in second (s))
    by CARD and SSM on CIFAR-100. $f_{\phi}$ is the pre-trained base classifier.}
    \label{tab:ACCandPA}
\end{table*}

\subsection{Comparison with GBM}
\label{ex:com:GBM}

In this subsection, we elaborate on the similarities and differences between SSM and GBM methods, including NGboost and DBT.

From an algorithmic perspective, obtaining the score function, i.e., log-likelihood estimation, is generally challenging, making it difficult to obtain the maximum likelihood estimate in traditional GBM, including NGboost. Accordingly, to solve this issue, our SSM approach utilizes denoising score matching for score estimation, which has been proven on a variety of tasks that better estimation results can be achieved.

From the perspective of network structure, we train a unified model for the denoising process, like most score-based generative models. Compared to using multiple weak learners in GBM, utilizing a single model for gradient boosting can achieve parameter sharing, helping the network learn more general and powerful feature representations throughout the denoising process, and improve training efficiency.

Moreover, compared to DBT, which is also a score-based gradient boosting model, SSM utilizes the inference process in the same way as the original score-based generative model instead of a decision tree structure. Although using a decision tree model can enhance the interpretability of the model, the output process of the model is more dependent on the number of leaves. When the number of leaves is large, i.e., the denoising timesteps are long, the DBT model based on the decision tree cannot be effectively accelerated using the acceleration algorithm, including DDIM, resulting in a relatively long inference time for the model, which is also one of the advantages of SSM.

Additionally, in the traditional GBM model, the weak learners only use the original input, which leads to an unclear relationship between different learners. By leveraging denoising score matching, the proposed SSM has two advantages:
\begin{itemize}
    \item The previous denoising result is always used as input in the next denoising steps, which ensures the Markov property and enhances the correlation in denoising process.
    \item Different noise levels ensure the focus of model learning and improve learning efficiency.
Moreover, the performance of GBM model is limited by the settings of the hyperparameter. 
\end{itemize}

In this paper, we discussed the setting of hyperparameters in detail and combined ablation experiments to ensure the effectiveness of our method in Section 4.

\subsection{Comparison with CARD}
\label{ex:com:card}

Lastly, we make a detailed comparison between SSM and CARD and provide an insight analysis on their differences.

First, SSM is a score-based gradient boosting model, and CARD is a diffusion-based model. Although the two models are somewhat related, differences in design motivations lead to differences in the training and inference processes. The score-based model aims to estimate the gradient of unknown data distribution via score matching. The added noises are to facilitate estimation, where more noises can ensure that the gradients of more points in the space can be trained. Additionally, via the Langevin dynamics, we can approximate an unknown distribution and generate data from that distribution \cite{song2019generative}. 
On the other hand, the diffusion-based model makes the unknown distribution converge to the standard normal distribution by adding noise. By learning the added noise, we can denoise the noise to the original distribution step by step, generating new data \cite{ho2020denoising}. 

This difference leads to the difference in the number denoising steps between the two methods. The score-based model needs an enough number of steps to make the Langevin dynamics converge. Since it estimates the noise added at each step, the diffusion-based model can only denoise according to the specified number of steps. Therefore, the proposed model SSM can utilize score function as the
current estimated error distance to ensure the convergence of the Langevin dynamics, thus reducing the number of denoising steps, while CARD cannot achieve the same result. 

\begin{figure*}[h]
    \centering
    \includegraphics[width= 0.95\linewidth]{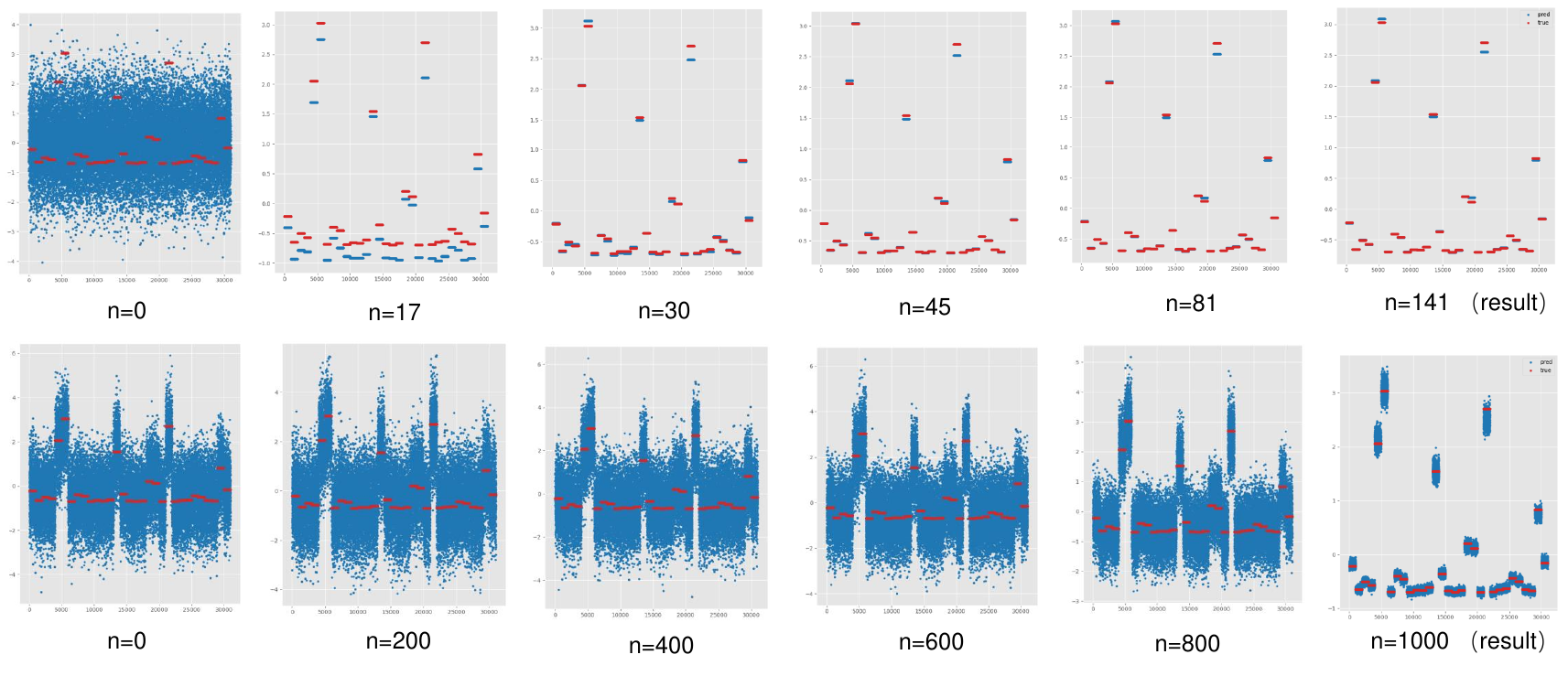}
    \caption{One batch prediction results corresponding to different denoising steps on UCI-Yacht task by SSM (top) and CARD (bottom). $n$ represents the number of denoising steps. The red and blue points represent the true values and the prediction results generated by 1000 samples respectively.}
    \label{fig:cardandssm}
\end{figure*}

Another important difference is the different goals of model prediction. Since SSM removes the noise term in the Langevin dynamics, each step of denoising moves towards the maximum likelihood estimation point, i.e., the prediction of the input data. Even if we start denoising from different initial points, the convergence will always be concentrated in a small range of the target.
Relatively, CARD and other Bayesian neural network frameworks are probabilistic prediction models, whose aim is to the distribution that the outcome may follow. To illustrate this clearly, 
the prediction results corresponding to different denoising steps on UCI-Yacht task by SSM and CARD are shown in Figure \ref{fig:cardandssm}. It is noticed that for 1000 prediction points generated by the same input, a more compact prediction result can be obtained by SSM, which also shows that SSM is a more deterministic model. 
Moreover, the SSM figure is plotted at the step when SSM switches to the next noise level for denoising. It is noticed that under the guarantee of error estimation, SSM can indeed ensure that the current state converges within the error range of the target point.

In summary, SSM is a relatively deterministic prediction model. Additionally, since the inference process is constantly shortening the distance between the current state and the target state, SSM can ensure the convergence of the model based on the error and shorten the inference time.

% PICP and QICE unsutiable

% and Negative log likelihood (NLL), which is a metric for uncertainty estimation,

\subsection{Limitations and Future Work}
\label{ex:lim}

In this section, we discuss the limitations of SSM and our future work. 

First, we apply Technique \ref{tech:11111} by removing the noise term in the Langevin dynamics to make the model deterministic. Even though SSM outperforms the probabilistic models including CARD on several UCI tasks and CIFAR datasets, its deterministic prediction ability makes it possible to perform poorly when dealing with some tasks where the test set is too different from the training set, such as out of distribution dataset \cite{yang2021generalized}. For this type of task, probabilistic models can handle it more effectively because they predict probability distributions rather than deterministic values \cite{ren2019likelihood}. Second, according to the average number of denoising steps shown in Table \ref{tab:numberUCI2}, we notice that the upper step number bound approximated by Technique \ref{tech:end-signal} is usually smaller than the actual value. This is because we do not consider the network error when computing the end-signals, which may lead to a large gap between the estimated number of steps and the actual number, making it impossible to guarantee the completion time in practical applications. Additionally, due to the network error, the condition in Alg. \ref{alg:2} line 6 cannot always be satisfied, which will cause the current number of denoising steps to reach the given limited number. As a reuslt, there is no guarantee of error bounds in the subsequent denoising processes, leading to a decrease in prediction accuracy. Finally, We use the average loss of the network on the training set as the error of the score network used in the Technique \ref{tech:three}, which may also lead to an inaccurate steps number approximation. Besides, when the lower bound computed by Technique \ref{tech:three} is small, this affects Techniques \ref{tech:lossssss} whose aim is to train more on the last noise level. 

In our future work, we plan to incorporate uncertain items into the SSM, such as in the last denoising stage, to enhance its ability to predict the probability distribution of the target value instead of the target value itself. Moreover, we will consider more accurate estimates from the previous period to minimize the impact of network errors on the estimation results. Finally, Our estimates of inference time and error are from a numerical perspective rather than a distributional perspective, which are not applicable to the original score-based generative model. Therefore, we will consider modifying our technique to accommodate score-based generative models.

% \todo{gu ji de shi ji yu wucha de}

% \todo{yu gai lv mo xing de chayi}

% \todo{yi lai yu yu xun lian de }

% % \todo{ping jun bu chang de jieguo}

% \todo{yu dddim bijiao cifar-10}

\end{document}